\documentclass{article}

\usepackage{arxiv}

\usepackage[utf8]{inputenc} % allow utf-8 input
\usepackage[T1]{fontenc}    % use 8-bit T1 fonts
\usepackage{hyperref}       % hyperlinks
\usepackage{url}            % simple URL typesetting
\usepackage{booktabs}       % professional-quality tables
\usepackage{amsfonts}       % blackboard math symbols
\usepackage{nicefrac}       % compact symbols for 1/2, etc.
\usepackage{microtype}      % microtypography

\usepackage[pdftex]{graphicx}

\usepackage{amsmath}
\usepackage{amssymb}
\usepackage{algorithm}
\usepackage[noend]{algpseudocode}
\usepackage{xcolor}
\makeatletter
\def\BState{\State\hskip-\ALG@thistlm}
\makeatother
\DeclareMathOperator*{\argmin}{arg\,min}

\title{Autonomous Identification and Goal-Directed Invocation of Event-Predictive Behavioral Primitives}

\author{
  Christian ~Gumbsch\\
  Max Planck Institute for Intelligent Systems\\
  \& University of T\"ubingen\\
   T\"ubingen, Germany \\
  \texttt{christian.gumbsch@tuebingen.mpg.de} \\
   \And
 Martin V.~Butz \\
  University of T\"ubingen\\
   T\"ubingen, Germany \\
  \texttt{martin.butz@uni-tuebingen.de} \\
     \And
 Georg Martius \\
  Max Planck Institute for Intelligent Systems\\
   T\"ubingen, Germany \\
  \texttt{georg.martius@tuebingen.mpg.de} \\
}

\begin{document}
\maketitle

\begin{abstract}
  Voluntary behavior of humans appears to be composed of small, elementary building blocks or behavioral primitives. While this modular organization seems crucial for the learning of complex motor skills and the flexible adaption of behavior to new circumstances, the problem of learning meaningful, compositional abstractions from sensorimotor experiences remains an open challenge.
  Here, we introduce a computational learning architecture, termed surprise-based behavioral modularization into event-predictive structures (SUBMODES), that explores behavior and identifies the underlying behavioral units completely from scratch.
  The SUBMODES architecture bootstraps sensorimotor exploration using a self-organizing neural controller. While exploring the behavioral capabilities of its own body, the system learns modular structures that predict the sensorimotor dynamics and generate the associated behavior.
  In line with recent theories of event perception, the system uses unexpected prediction error signals, i.e., surprise, to detect transitions between successive behavioral primitives.
  We show that, when applied to two robotic systems with completely different body kinematics, the system manages to learn a variety of complex and realistic behavioral primitives. Moreover, after initial self-exploration the system can use its learned predictive models progressively more effectively for invoking model predictive planning and goal-directed control in different tasks and environments.
\end{abstract}


\section{Introduction}
Opening the fridge, grasping the milk and  drinking from the bottle -- behavioral sequences, composed of multiple, smaller units of behavior, are ubiquitous in our minds \cite{Butz:2017,Flash:2005,Gaerdenfors:2014}.
More generally speaking, we  humans seem to organize our behavior and the accompanying perception into small, compositional structures in a highly systematic manner \cite{Butz:2016}.
These structures are often referred to as \textbf{building blocks} of behavior or \textbf{behavioral primitives} and can be viewed as elementary units of behavior above the level of single motor commands \cite{bentivegnaPrimitives}.

A large challenge for the brain as well as artificial cognitive systems lies in the effective segmentation of our continuous perceptual stream of sensorimotor information into such behavioral primitives.
When does a particular behavior commence?
When does it end?
How are individual behavioral primitives encoded compactly?
In most cognitive systems approaches so far, behavioral primitives are segmented by hand, pre-programmed into the system, or learned by demonstration \cite{schaalMovementPrimitives,Ijspeert:2013,Nguyen-Tuong:2011,Worgotter:2013}.
In all cases, though, the primitives are made explicit to the system, that is, the learning system does not need to identify the primitives autonomously.
Our brain, however, seems to identify such primitives on its own, starting with bodily self-exploration.

Here, we introduce a computational architecture, termed SUrprise-based Behavioral MODularization into Event-predictive  Structures (SUBMODES), that learns behavioral primitives as well as behavioral transitions completely from scratch.
The SUBMODES architecture learns such primitives by exploring the behavioral repertoire of an embodied agent.
Initial exploration is realized by a closed loop control scheme that adapts quickly to the sensorimotor feedback.
In particular, we use differential extrinsic plasticity (DEP) \cite{dep1}, which causes the agent to explore body-motor-environment interaction dynamics.
DEP essentially fosters the exploration of coordinated, rhythmical sensorimotor patterns, including a tendency to `zoom' into particular dynamic attractors, stay and explore them for a while, and upon small perturbations leave one attractor in favor of another one.
Starting with this self-exploration mechanism, the algorithm learns internal models that are trained to predict the motor commands and the resulting sensory consequences of the currently performed
behavior.

The SUBMODES system uses an unexpected increase in prediction error to detect the transition from one behavioral primitive to another.
If such a `surprising' error signal is perceived, the internal predictive model either switches to a previously learned model or a new model is generated if the behavior was never experienced before.
In this way, the agent systematically structures its perceived continuous stream of sensorimotor information online into modular, compositional models of behavioral primitives as well as predictive event-transition models.
We show that a large variety of behavioral primitives can be learned form scratch even in robotic systems that have both many degrees of freedom and interact with complex, noisy environments.
Moreover, we show that after initial self-exploration the agent can use its learned predictive models progressively more effectively for invoking goal-directed planning and control.
In effect, the system learns predictive behavioral primitives and event transition models to invoke hierarchical, model-predictive planning \cite{Barto:2003,Botvinick:2014}, anticipating the sensory consequences of the available behaviors and choosing those behavioral primitives that are believed to bring the system closer to a desired goal state.

In sum, the main contributions of this work are as follows:
(i) we show how a self-organizing behavior control principle can be utilized to systematically explore the sensorimotor abilities of embodied agents;
(ii) we introduce an online event segmentation mechanism, which automatically structures the generated sensorimotor experiences into predictive behavioral and event-transition encodings;
(iii) we show how such encodings can be used for hierarchical planning and goal-directed behavioral control.
We evaluate the novel techniques in complex, simulated robots that are acting in noisy, physics-based environments.

\section{System Motivation and Related Work}
The problem of abstracting our sensorimotor experiences into conceptual, compositionally meaningfully re-combinable units of thought is a long-standing challenge in cognitive science, including cognitive linguistics, cognitive robotics, and neuroscience-inspired models \cite{Barto:2003,Butz:2017,Gaerdenfors:2014,Giese:2003,Herbort:2005,Sugita:2011a,Tani:1999}.
One important type of such units concerns concrete behavioral interactions with the environment, regardless if they lead to transitive motions of the body or of other objects.
Depending on the level of abstraction and the field of research, different synonyms can be found in the literature \cite{Flash:2005}, such as \lq behavioral primitives\rq \space \cite{bentivegnaPrimitives},  \lq movement primitives\rq \space \cite{schaalMovementPrimitives}, \lq motor primitives\rq \space \cite{Ijspeert:2013},
\lq motor schemas\rq \space \cite{motorSchema}, or \lq movemes\rq \cite{movemes}.
It has been suggested that our ability to serially combine these compositional elements is crucial for our ability to quickly learn complex motor skills and to flexibly adjust our behavior to new tasks \cite{schaalMovementPrimitives}.
Furthermore, the assumption that there exists a limited repertoire of behavior, has been proposed as a way to deal with the curse of dimensionality and redundancy at different levels of the motor hierarchy, moving from simple behavioral primitives towards an ontology of more sophisticated interaction complexes \cite{Flash:2005,Kraft:2008,Nguyen-Tuong:2011,Worgotter:2013}.

Although the acquisition and application of behavioral primitives has been extensively studied in cognitive robotics and related fields, it is still not clear how we discover, encode, and ultimately use these behavioral primitives for the effective invocation of goal-directed behavioral control.

The remainder of this section is structured as follows:
In Section~\ref{relatedWorkA} we introduce cognitive and computational theories on how goal-directed behavioral control is learned by both humans and artificial systems.
In Section~\ref{relatedWorkB} we provide an overview on how continuous sensorimotor information can be converted into compositional, temporally predictive encodings of behavior.
Finally, in Section~\ref{relatedWorkC} we outline how these compositional abstractions can guide higher-order, hierarchical planning.

\subsection{Goal-directed behavioral control} \label{relatedWorkA}
According to the Ideo-Motor Principle \cite{james, ABC, stock}, encodings of behavior are closely linked to their sensory effects.
The main idea is that initially purely reflex-like actions are paired with the sensory effects they cause.
 At a later point in time, when the previously learned effects become desirable, the behavior can be applied again \cite{stock,Butz:2017}.
While the Ideo-Motor Principle was heavily criticized and ridiculed during the beginning of the 19th century and in the era of Behaviorism, it has seen a revival over the last decades in various fields of cognitive science, as, for example, manifested in the propositions of the Anticipatory Behavioral Control (ABC) theory \cite{ABC} as well as the Theory of Event Coding (TEC) \cite{Prinz:1990}.

TEC suggests that perceptual information and action plans are encoded in a common representation.
According to TEC, actions and their consequent perceptual effects are encoded in a common predictive network, which allows the anticipation of perceptual action consequences and the inverse, goal-directed invocation of the associated motor commands.
TEC implies that behavior is primarily learned with respect to the effects that it produces.
The ABC theory focuses even more on the learning of sensorimotor structures.
According to ABC, the critical conditions for the application of an action-effect encoding are learned by focusing on (unexpected) perceptual changes, which lead to a further differentiation of conditional structures \cite{Butz:2002c}.
For example, it can be learned that an object first needs to be in reach before we are able to grasp it \cite{Butz:2017,butzStructures}.
In sum, both theories emphasize that our brain encodes behavior with respect to the effect it entails and it does so, because the resulting structures enable the selective and highly flexible activation of an action-effect complex depending on the current context and desired goal states.

Along similar lines, Wolpert and Kawato have proposed that our brain may learn modular forward-inverse model pairs to acquire progressively more complex motor skills \cite{wolpert1}.
The proposition was implemented later on in the MOSAIC system \cite{wolpert2}.
The MOSAIC system learns sets of discrete, internal models, each consisting of a forward model, which predicts the sensory consequence of an action, and a paired inverse model, which generates the required motor commands.
For each internal model, the forward model is used to determine which behavior is most likely responsible for the observed sensory dynamics, while the inverse model can generate the associated motor commands.

The learning of behavioral control has also been examined within the reinforcement learning (RL) framework \cite{SuttonBarto2018:RLIntro}.
In RL one particular control policy is trained to maximize given rewards.
Under appropriate conditions, such a policy can correspond to a particular behavioral primitive trained on a specific task.
The learning and task-dependent optimization of movement primitives has, for example, been investigated
in an Actor-Critic framework \cite{schaalMovementPrimitives}.
It has been shown that complex movement primitives, in realistic settings, such as \lq hitting a baseball with a bat\rq \space can achieve nearly optimal performance when applying policy gradient based optimization \cite{naturalActorCritic}.
Various alternative approaches have been investigated and contrasted \cite{Calinon:2009,Ijspeert:2013,Kober:2011,Nguyen-Tuong:2011,Sigaud:2011}.
In all cases, the beginning and end of a movement primitive is predefined and not autonomously discovered by the system itself.
Furthermore, classical model-free RL methods typically require much more time to learn complex behavioral dynamics than predictive, model-based approaches.

\subsection{Learning sensorimotor abstractions} \label{relatedWorkB}
While the outlined theories give an account on how behavior can be encoded, they do not explain how the continuous stream of sensorimotor information may be structured systematically to infer the underlying behavioral primitives.
Event segmentation theory (EST) \cite{zacks} gives a concrete formulation of how our brain might be able to segment the perceptual stream into discrete representations.
According to EST, humans perceive activity in terms of discrete conceptual events.
An event is defined as ``a segment of time at a given location that is conceived by an observer to have a beginning and an end'' \cite[p. 3]{zacksTversky}.
This definition of an event is rather general, containing both short sensorimotor events, such as  \lq grasping a mug\rq , but also potentially long segments with multiple agents and ongoing activities, e.g., a concert.
When considering the learning of behavioral primitives, we can focus solely on the individual sensorimotor level of events.

According to EST, our perceptual process is guided by a set of internal models, which continuously predict what is perceived next.
A specific set of event models is active over the course of one event, i.e., until a transient increase in prediction error occurs.
Such a transient error signal may result in a change in the currently active internal models.
EST further suggests, that such a prediction error-based segmentation mechanism might occur on different levels of abstraction, resulting in a hierarchical, taxonomic organization of events \cite{zacks, zacksTversky}.
Hence, according to EST a cognitively plausible way to conceptualize the continuous sensorimotor stream into compositional behavioral models is based on transient error signals of internal predictive models -- essentially a more concrete formalism that dovetails with the ABC theory.

Segmentation mechanisms based on transient prediction error signals have been studied in various computational models:
Predicting movements in video sequences of actors performing everyday motions, paired with the dedicated processing of transient prediction error signals, led to the discovery and encoding of simple movement primitives in a recurrent neural network \cite{zacksComputationalModel}.
Similarly, learning predictive models and using an unexpected increase in prediction error has been used to learn forward models of different object interaction events in simple, physics-based simulation environments \cite{gumbsch1, gumbsch2}.
In both systems, the prediction error-based detection mechanism works online.
The basic principle can be closely related to a surprise-based perceptual processing mechanism, which has been shown to segment a hierarchically structured environment (four-rooms problem) into its sub-components (individual rooms) even in the case of very high noise \cite{surprise}.

Related mechanisms that use perceptual prediction errors or prediction confidence to gate the learning signal while learning different types of behavior have been applied in various control systems \cite{wolpert2, mixtureExperts, Butz:2018, butz2018learning, tani2003, murata2014learning}.
Mechanisms that focus on learning progress or more graph-based algorithms to detect transitions have been proposed as well \cite{duminy2018learning, Schapiro:2013,Simsek:2004,Simsek:2009}.

\subsection{Planning based on hierarchical structures}
\label{relatedWorkC}

Learning temporal abstractions of behavior enormously simplifies goal-directed planning in high-dimensional systems.
If the right behavioral primitives are available rather complex tasks, such as \lq drinking from a mug\rq, can be decomposed into a sequence of primitives (\lq reaching\rq , \lq grasping\rq , \lq lifting\rq , etc.).
This drastically reduces the search space for planning and control \cite{bentivegnaPrimitives, Flash:2005, duminy2018learning}:
Instead of choosing a motor command from the entire space of possible motor actions, once the next primitive is identified, a much smaller subspace of actions can be analyzed to determine the next motor command.

From a predictive coding-inspired, neuro-robotics perspective, hierarchical behavioral planning was, for instance, imlemented in a recurrent neural network architecture \cite{tani2003}.
A two level hierarchy is employed where the levels interact in a bottom-up and top-down manner:
The higher level produces top-down expectations of the ongoing behavior, essentially encoding sequences of behavioral primitives.
The lower level produces sensorimotor predictions based on the perceptual input and the top-down estimations.
Prediction errors from the low level are, in turn, used to update activity of the high level in a bottom-up fashion.
Related approaches integrate multiple time-scales for the adaption within the different levels of the hierarchy \cite{tani2008, tani2017, zhong2018}.

Discovering behavioral primitives and applying them for high-level goal-directed control is closely related to hierarchical RL and the options framework \cite{Barto:2003, Botvinick:2009,Sutton:1999}.
An option is defined as a ``generalization of primitive actions to include temporally extended courses of action'' (\cite{Sutton:1999}, p. 186).
In the right setting, i.e., an embodied, robotic agent with an elementary action corresponding to a single motor command, an option can resemble both a behavioral primitive or a series of behavioral primitives, e.g. `grasping an object'.
In the options framework a particular option is typically defined with respect to a specific subgoal state.
For example, the `grasping an object'-option might terminate when the object is held by the hand of the agent.
An option can then be trained by comparing the outcome of performing the option with the desired subgoal to determine a pseudo-reward and updating the internal structures reward-dependently \cite{Botvinick:2009}.
While recent implementations of hierarchical deep RL have shown remarkable performance in rather challenging video gaming tasks \cite{tenenbaumMontezuma}, self-motivated behavioral exploration and effective subgoal identification remain as open challenges.

\begin{figure*}
\begin{center}

\includegraphics[scale=0.11]{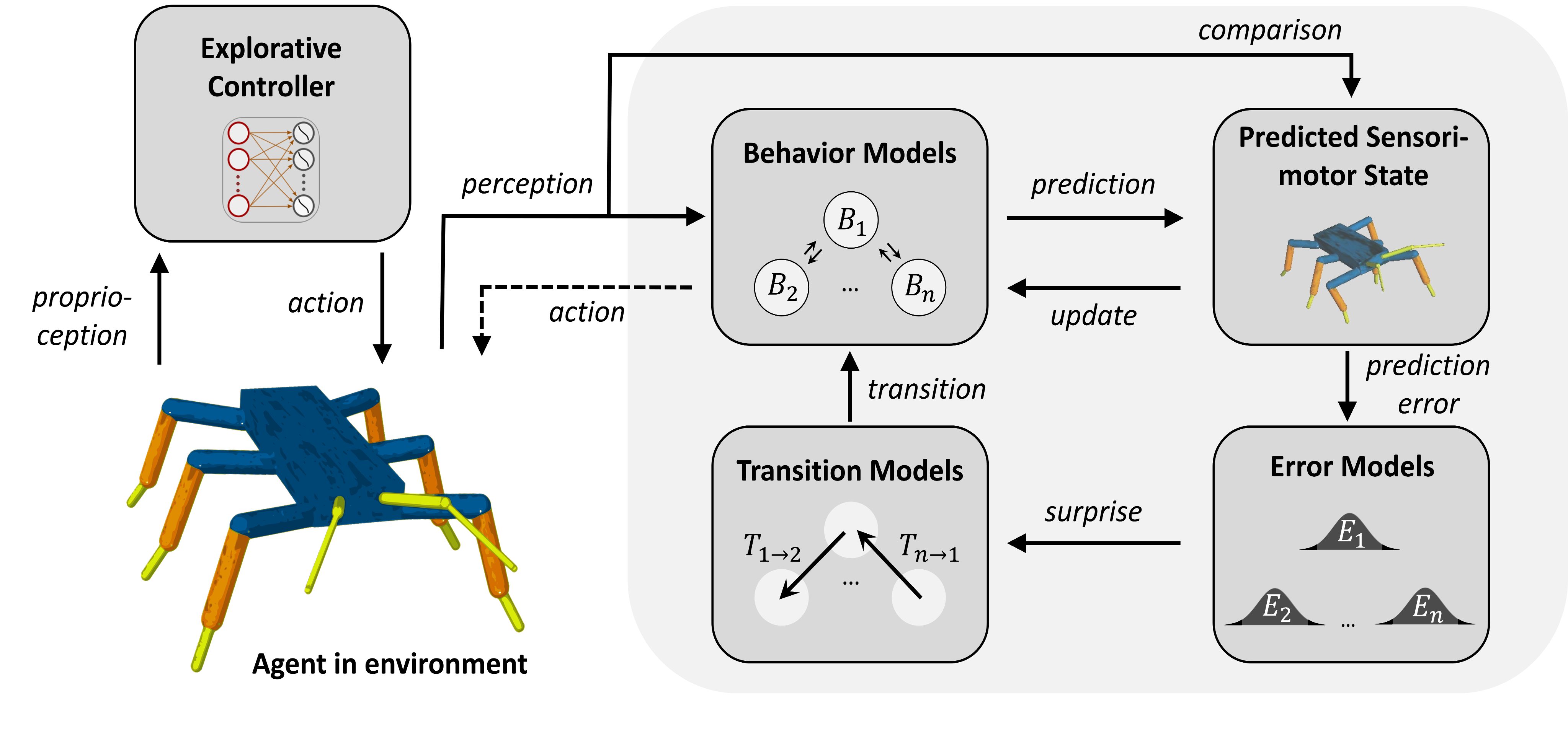}

\caption{Illustration of the SUBMODES architecture during the learning of behavior. An \emph{explorative controller} generates motor commands based on the current proprioceptive input to explore self-organizing behavior. One of multiple, internal \emph{behavioral models} attempts to predict the motor commands and sensory consequences of the ongoing behavior. The \emph{predicted sensorimotor state} is compared to the actual state to compute the prediction error and update the active behavioral model. For each behavioral model an \emph{error model} is trained, estimating the prediction confidence. If surprise is detected, i.e., a strong error signal outside the usual prediction confidence, the system is allowed to exchange the active behavioral model.
For each transition between two different behavioral models a \emph{transition model} is learned.
During goal-directed control, the explorative controller is deactivated and the active behavioral model determines the next action (dashed line). \label{system}}
\end{center}
\end{figure*}

\section{Overview of the SUBMODES architecture}
\label{Submodes}

We propose a computational architecture, termed SUrprise-based Behavioral MODularization into Event-predictive Structures (SUBMODES), to discover behavioral primitives and learn event-predictive models of the corresponding behavior for an embodied agent completely from scratch.
The SUBMODES architecture uses different modular components to explore and learn behavioral primitives and detect transitions in behavior,  illustrated in Fig.~\ref{system}.
In this section we give an overview of the system.
In the Appendices \ref{sectionDEP} -- \ref{sectionParameters} further algorithmic details are provided.
In the Appendix \ref{sectionPseudocode} the system is described in terms of pseudocode.

The SUBMODES architecture is composed of different modular components, responsible for exploring behavior, learning models for different behavioral modes, and detecting and encoding transitions in behavior.
The different behavioral primitives learned by the system are encoded in \textbf{behavioral models} of our learning architecture.
These models receive sensorimotor perceptions about the agent as an input and produce a \textbf{predicted sensorimotor state}, anticipating future sensorimotor perceptions and actions.
We assume that the system switches between its behavioral modes in a predictable fashion, whereby the occurrence of such transitions is detected by \textbf{error models}.
Upon detecting a transition, \textbf{transition models} are trained to encode the critical conditions that enable such a change in behavior and the sensory consequences thereof.
Initially, behavioral exploration is bootstrapped by an \textbf{explorative controller} and the behavioral models are trained on the perceived sensorimotor experiences.
At a later phase, the explorative controller is deactivated and the system can use its learned representations of behavior for anticipatory goal-directed control.

The SUBMODES system learns behavioral primitives based on the experienced sensorimotor time series.
We bootstrap this learning process by invoking motor commands via a neural network controller that is updated using differential extrinsic plasticity (DEP) \cite{dep1}.
At every discrete time step $t$ the controller transforms proprioceptive sensor values $x(t) = (x_1, x_2, ..., x_n)$ into motor commands $y(t) = (y_1, y_2, ..., y_m)$.
Here, we use a one-layered feed-forward neural network, as
\begin{equation}
y_i(t) = \tanh \left( \sum_{j=1}^{n} W_{ij} x_j(t) + h_i \right), \label{equCNetwork}
\end{equation}
for a motor neuron $i$, with $W_{ij}$ the weight connecting input $j$ with the output neuron $i$ and a bias term $h_i$.

With fixed weights $W$ the controller would continuously generate motor commands corresponding to one particular behavioral pattern.
However, the network weights $W_{ij}$ are constantly changed by applying the DEP-learning rule.
This learning rule essentially updates the weights based on correlations of sensoric velocities over some time $\phi$, i.e.,
\begin{equation}
\Delta W(t) \propto M\dot{x}(t- \phi)\dot{x}(t),
\end{equation}
with $M$ an inverse model describing the relationship between motor actions and proprioceptive sensor values (details in Appendix \ref{sectionDEP}). Besides the weight updates, changes in behavior can also arise from a \textit{bias dynamics}, which after some time of inactivation shifts the bias value $h_i$ for the most inactive motor neurons $i$.

When applying the explorative controller using the DEP learning rule to an embodied agent, the controller typically discovers different dynamic sensorimotor attractors, which correspond to behavioral dynamics that unfold relatively uniformly over time.
These behavioral dynamics can be seen as behavioral primitives, since they typically correspond to simple elementary actions like \lq crawling\rq , \lq shaking hands\rq \space or \lq wiping a table\rq \space \cite{dep2}.
However, upon perturbations the controller might leave one sensorimotor attractor and some time later discovers a new one, resulting in a change in behavior.
Such perturbations can be caused by a sudden change in interaction of the agent with its environment, e.g., by hitting an obstacle, or by changes within the sensorimotor loop, e.g., the activation of a bias neuron.
This property makes the DEP-controller an ideal candidate for behavioral exploration of a complex, embodied agent.

The SUBMODES architecture encodes the explored behavioral primitives through a set of modular, predictive \emph{behavioral models} $\mathcal{B}$.
One behavioral model $B_i \in \mathcal{B}$ attempts to encode one particular behavioral primitive previously demonstrated by the explorative controller.
Each model $B_i$ is a single-layered neural network (no hidden layer) receiving the current sensory state $x(t)$ as an input and predicting the next motor command $y'(t)$ and the sensory consequence of this particular action $\Delta x'(t+1)$.
At a certain point in time $t$ only one model $B(t) = B_i$ is active.
The sensorimotor predictions produced by the active model $B(t)$ are compared to the perceived change in sensory values $\Delta x(t+1)$ and motor command $y(t)$ and the prediction error is computed as the deviation between prediction and sensation. The error signal is then used to update the active model $B_i$ using delta-rule based gradient descent.

To maintain minimal statistics about the accuracy of the sensory predictions, the system contains a set of \emph{error models} $\mathcal{E}$.
For each behavioral model $B_i$ an error distribution $E_i \in \mathcal{E}$ is learned, which is estimated by means of a normal distribution.
Each error model $E_i$ maintains a moving average $\bar{e}_i(t)$ and variance $\bar{\sigma}_i(t)$ of the sensory prediction error, thus, estimating the first two moments of the prediction for each behavioral model.

We assume that changes in behavior result in a strong, unexpected increase in the sensory prediction error $e(t)$ for the currently predicting model.
The system detects such a \emph{surprise} \space for time step $t$ if
 the error is outside of a certain confidence region of the error statistics,
\begin{equation}
e(t) > \bar{e}_i(t) + \theta \bar{\sigma}_i(t), \label{surprise}
\end{equation}
with $e(t)$ the current sensory prediction error\footnote{In practice, we compute $e(t)$ over a short time frame of 25 time steps.}, $\bar{e}_i(t)$ the moving average and $\bar{\sigma}_i(t)$ the moving error deviation of the currently active behavioral model $B_i$ and $\theta$ the threshold~\cite{surprise}.

If a surprise signal is detected, the system is allowed to switch its active behavioral model $B(t)$.
To determine the new model, the system enters a \textit{searching period}.
In this mode, the mean prediction error of all existing models are monitored and if
 there is one which shows a non-surprising error (determined by Equation \ref{surprise}), this model takes over.
If after a maximum amount of time steps, the mean prediction error of every model is considered surprising a new model $B_j$ is generated and added to $\mathcal{B}$.
In this way, the system is able to switch between previously learned behavioral models and to generate new models on the fly.

While transitions in behavior are initially detected based on strong increases in prediction error, we assume that the system switches predictably between such behavioral primitives.
For example, some transitions in behavior may only occur in a specific context, for instance, a transition from walking to swimming may only occur in shallow water.
To model the critical conditions leading to a transition in behavior and, thus, enable the system to accurately predict such a transition, we train a set of transition models $\mathcal{T}$.
For each transition from model $B_i$ to model $B_j$ a transition model $T_{i \rightarrow j} \in \mathcal{T}$ is trained.
One transition model $T_{i \rightarrow j}$ attempts to identify the sensory state that allows this particular transition in behavior to take place and learns to predict how such a transition typically unfolds.
Transition models are updated once a transition in behavior occurs (further described in Appendix \ref{sectionTransitions}).
Hence, by learning models of transition in behavior, the SUBMODES architecture does not only learn how one stable behavioral primitive unfolds -- encoded by its behavioral models $\mathcal{B}$ -- but also how different behavioral primitives are connected through transitions in behavior -- encoded through transition models $\mathcal{T}$.

After an initial exploring and learning of behavioral abilities, the SUBMODES architecture can perform model-predictive planning to generate goal-directed behavior.
The predictive design of the internal models allows the system to directly use its learned structures for goal-directed control by minimizing the difference between anticipated and desired perceptions.
For goal-directed behavioral control, the motor command $y(t)$ is determined directly by the active behavioral model $B(t)$.
To plan behavior the system receives a desired sensory goal state $x^G(t)$ at every time step $t$.
The system first considers which subset of behavioral models $\mathcal{B}(t) \subseteq \mathcal{B}$ are applicable given the current sensory state using its transition models $\mathcal{T}$.
Then, the system \lq imagines\rq \space how the sensorimotor time series will unfold for each applicable behavior $B_j \in \mathcal{B}(t)$ over a fixed time horizon (details in Appendix \ref{sectionPlanning}).
By comparing the predicted time series with the goal state,  the system can activate the behavioral model whose predictions are closest to the goal state.

\begin{figure}
\begin{center}

\includegraphics[scale=0.22]{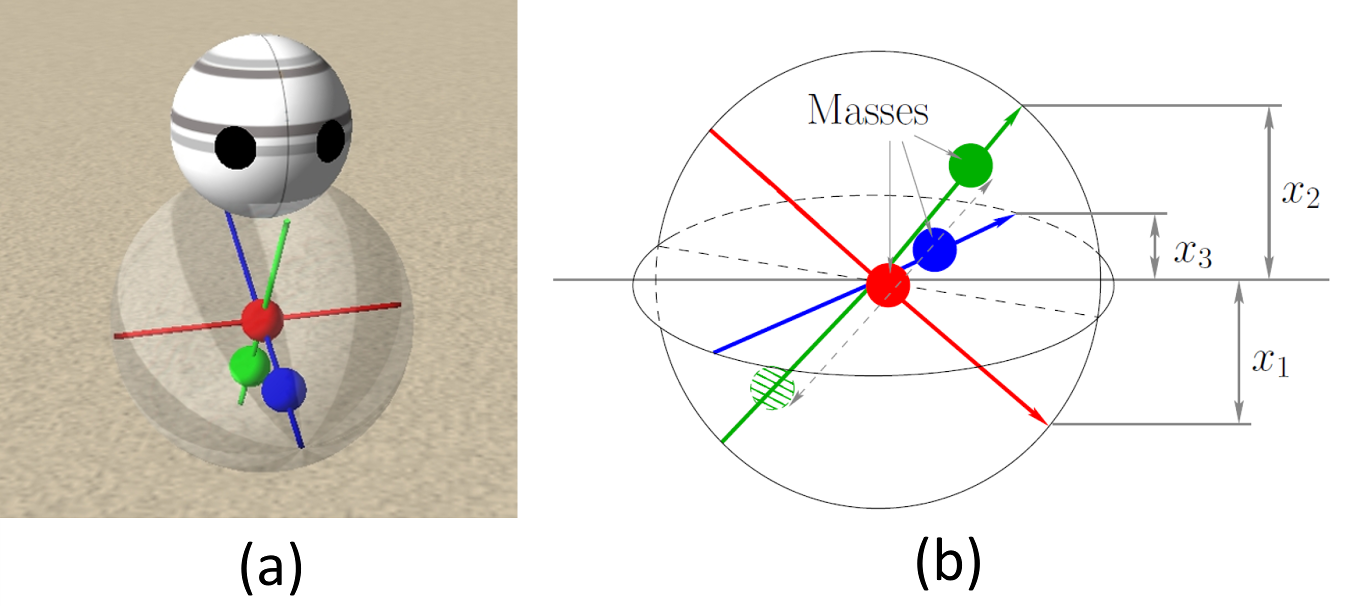}

\end{center}
\caption{Spherical robot and its axis orientation sensors. (a) shows a screenshot from simulation. (b) shows a schematic illustration of how the axis orientation sensor values $x_i$ are determined (taken from \cite{martiusPhd})  \label{spherical}}
\end{figure}

\section{Simulations}
The experiments were conducted in the physically realistic rigid body simulator \textsc{LPZRobots} \cite{lpzrobots}.
We tested the SUBMODES system on two robots, the \textit{Spherical robot} and the \textit{Hexapod}.
The system was updated with a frequency of 50\,Hz, each time receiving new sensor readings and setting motor commands.

The Spherical robot, illustrated in Fig. \ref{spherical}, has a ball shaped body, that contains three internal masses. The actuators move the masses along the axes where the target locations are specified by the motor commands:
$0$ corresponding to a centered position and $+1$ or $-1$ to the outer positions.
As  sensory information the projection of the axes' direction onto the $z$-component of the world-coordinate system is available, illustrated in Fig. \ref{spherical} (b).
The robot is equipped with a spherical head atop of its body to visualize the current rolling direction of the Spherical robot.
There is no physically interaction between the head and the body.
The head always \lq hovers\rq \space above the body and is rotating around its $z$-axis to face the current rolling direction.

The Hexapod is a six-legged robot inspired by a stick insect.
It has 18 actuated degrees of freedom, 3 in each leg.
Like in real stick insects, each leg is partitioned into three parts: femur, tibia, and tarsus.
The femur is connected to the body by a two-dimensional coxa joint, which is able to perform forward-backward and upward-downward rotations of the leg with respect to the body.
Femur and tibia are connected by a one-dimensional knee joint, which is able to rotate the tibia upward or downward with respect to the femur.
The motor values correspond to nominal angles of the joint, where $-1$ is associated with the minimal joint angle and $+1$ with the maximal angle.
Tarsi and antennae are attached by spring joints and are not actuated.

For both robots, the SUBMODES system receives the current  proprioceptive sensory information as an input.
When using the Hexapod, the delayed sensor values of the 12 coxa joints, with a small temporal delay of $\delta = 8$ time steps, are additionally provided.
Besides the proprioceptive sensory information, the velocity of the robot's body movement $v$ and the current orientation $\alpha$ are available sensory input.
The orientation $\alpha$ is provided in the form of $\sin(\alpha)$ and $\cos(\alpha)$.
Gaussian distributed noise  is added to the proprioceptive sensor values ($\sigma = 0.05$) and motor commands (Spherical: $\sigma = 0.05$, Hexapod: $\sigma = 0.1$).

\begin{figure*}
\begin{center}
  \includegraphics[scale=0.23]{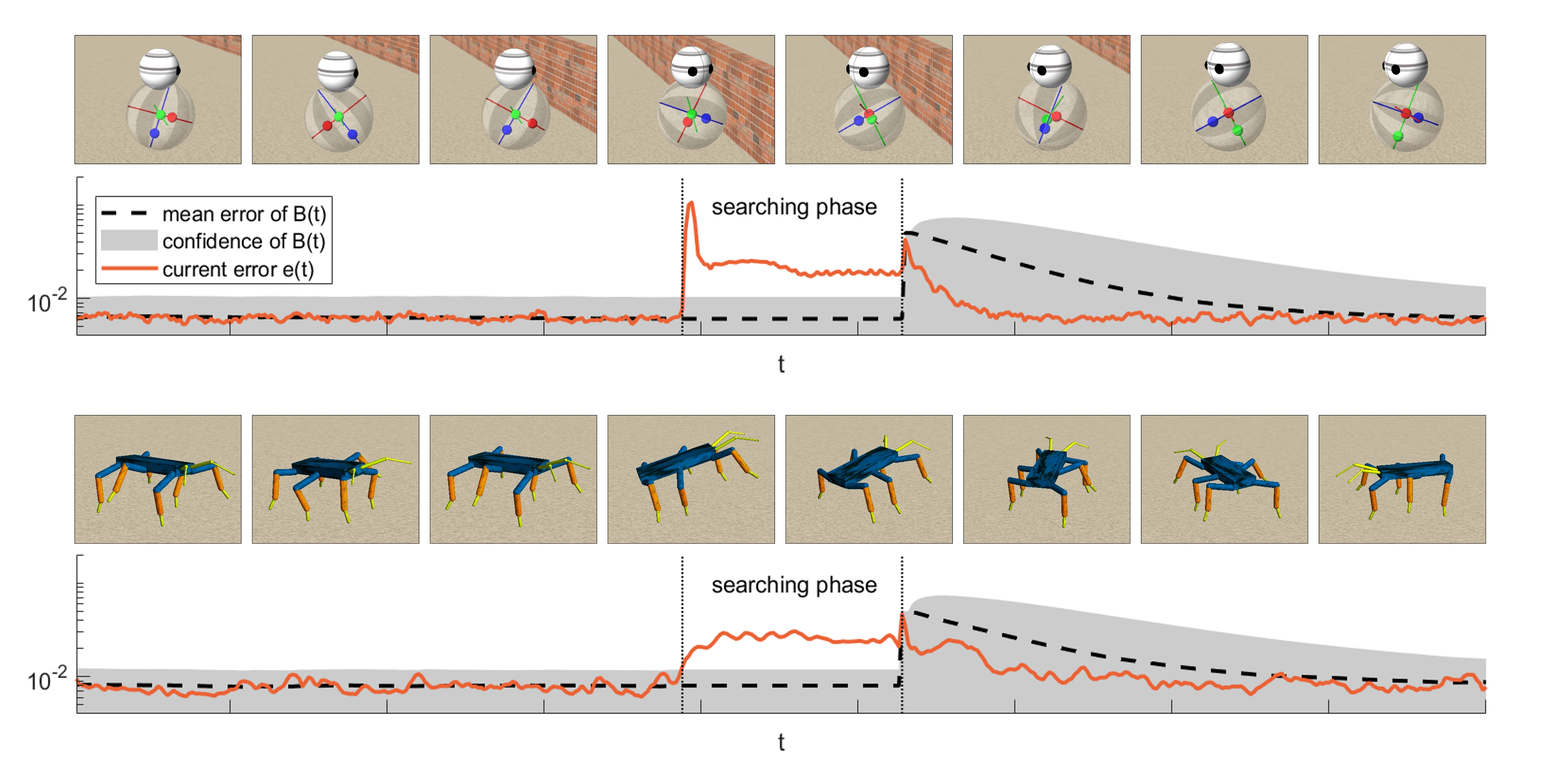}
\end{center}
\caption{ \label{surpriseFigure} Exemplary surprise detection for the Spherical robot and the Hexapod shown through the development of the internal error statistics over time.
The plots show the current prediction error ($e(t)$), the mean prediction error of the active model ($\bar{e}_i(t)$) and the confidence of the active model ($\bar{e}_i(t) + \theta \bar{\sigma}_i(t)$) over time.
Marks along the x-axis denote 10 second intervals.
The pictures show the surprise detection in simulation. The fourth frame depicts the time step when surprise was detected. The inter frame interval is approximately 0.5 seconds.
See the text for qualitative descriptions of the changes in behavior.}
\end{figure*}

\section{Results}

\subsection{Learned behavioral primitives}

In a first test, we examined which behavior is generated by the DEP-controller for the different robots, and how the SUBMODES architecture segments the explored stream of sensorimotor information into different behavioral primitives.
For that purpose, we let the SUBMODES system explore different behaviors for 90 minutes simulation time.

The Spherical robot was tested in a large quadratic arena surrounded by walls.
When applied to the Spherical robot, the DEP-controller typically generates different rolling motions, where one of its internal masses is kept fixed at the center of the respective axis, while the other two masses periodically oscillate with a certain phase shift.
Thereby, the robot's body rotates around one of its axis, while this axis is kept approximately parallel to the ground.
If the robot hits a wall, the sensorimotor dynamics are strongly perturbed.
These strong perturbations of the dynamics are amplified by the DEP-learning rule, which can result in the generation of a new rolling behavior.
If the robot continues one rolling motion long enough the bias dynamics, that we added to the original DEP-controller,  is activated and the previously centered weight is shifted to one side.
This results in a turning motion where the robot turns either left or right while rotating around the axis with the shifted internal mass.

In 90 minutes simulation time of exploring behavior for the Spherical robot, the SUBMODES system learned on average 15 behavioral models ($\sigma = 1.7$) over 10 simulations.
Surprise is typically detected by the system once the Spherical robot hits a wall or switches from rolling straight to driving a curve.
The upper part of Fig.~\ref{surpriseFigure} shows the detection of surprise for one exemplary transition in behavior.
In this example the robot first rolls in a straight line by rotating its body around its internal, green axis.
Upon hitting a wall the previously demonstrated behavior stops and for a short period of time all internal masses start moving.
This results in a strong increase in prediction error outside the confidence of the active model $B(t)$.
After some time the motion of one of the internal masses decreases (red mass) until this mass stops moving and is kept fixed at the center of the axis.
Since this behavior was demonstrated for the first time, no new model is found during the searching period and a new model is generated.
While the system performs the new rolling behavior, the predictions of this new behavioral model improve and the prediction confidence of this model decreases.
Further transitions in behavior are shown in Video 1 (\url{youtu.be/DKblfeM2Jys}).

\begin{figure}
\begin{center}

\includegraphics[scale=0.15]{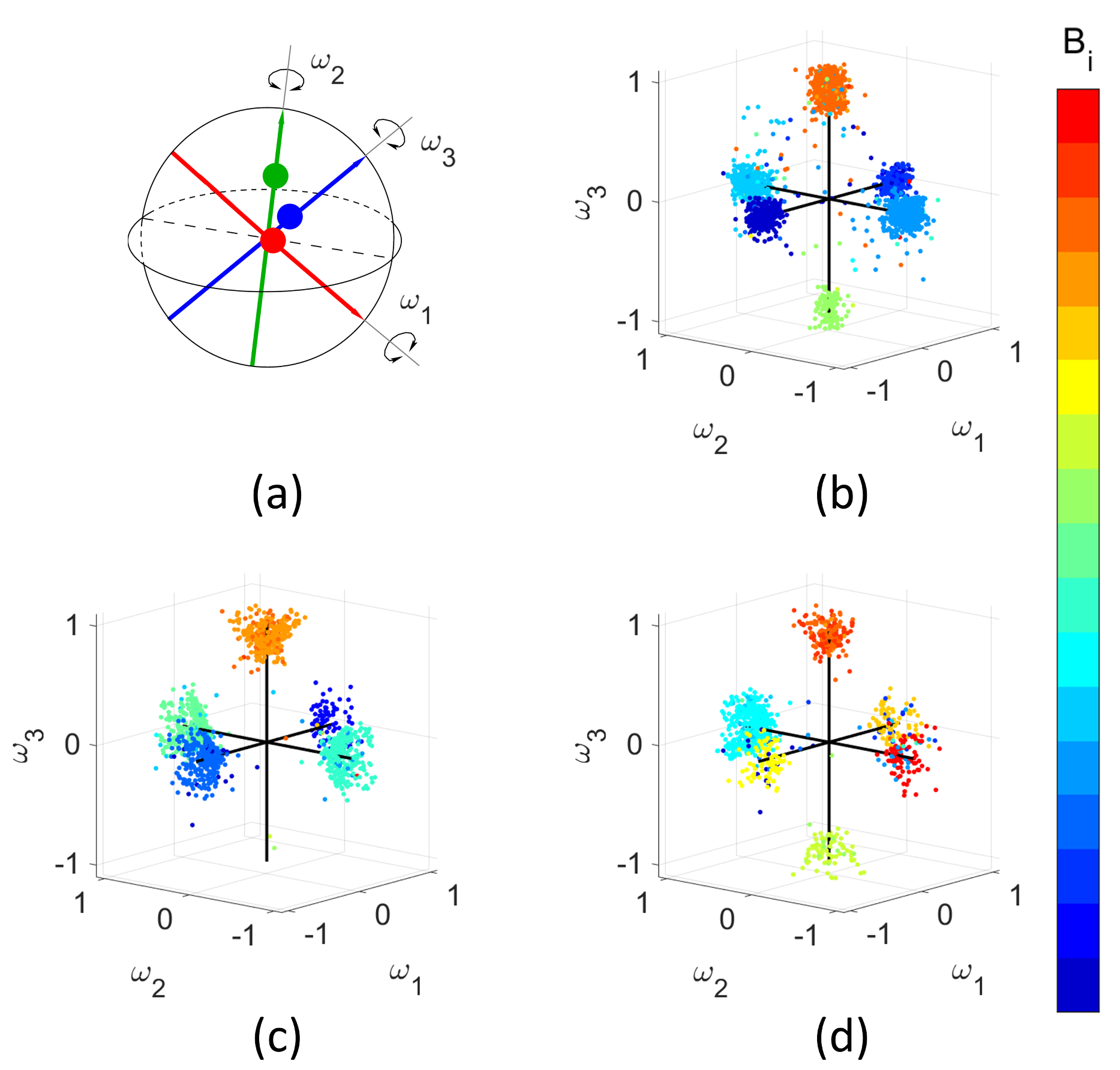}

\end{center}
\caption{\label{scatter} Behavioral space of the Spherical robot discovered by the SUBMODES architecture in one simulation. (a) illustrates the angular velocity $\omega_i$ around the internal axes. Each point in (b)-(d) shows the behavior  of the robot in terms of angular velocities $\omega_i$ at that time. (b) shows the behavior for rolling in an approximate straight line, i.e., with changes in driving direction $|\dot{\alpha}| < 0.3^\circ$. (c) shows the behavior for turning left ($\dot{\alpha} > 0.3^\circ$) and (d) shows the behavior for turning right ($\dot{\alpha} < -0.3^\circ$). The color of each point depicts which behavioral model $B_i$ was active and predicting the behavior at this time. For clarity only every 50th time step of the simulation is shown.}
\end{figure}

The behavior explored by the SUBMODES system for the Spherical robot can be described in terms of angular velocity $\omega_i$ for each axis $i$.
The angular velocity $\omega_i$ states how fast the body of the robot rotates around the internal axis $i$, illustrated in Fig. \ref{scatter} (a).
Fig. \ref{scatter} (b)-(d) depict rolling behaviors of the Spherical robot from one simulation in terms of angular velocities.
Since the change of orientation $\dot{\alpha}$ is not reflected in $\omega$, we separate the behavior for driving straight (Fig. \ref{scatter} (b)), driving a right curve (Fig. \ref{scatter} (c)), and driving a left curve (Fig. \ref{scatter} (d)).
Curved rolling corresponds to rotating around axis $i$, where the mass of axis $i$ is shifted to the right or left side of axis $i$.
The color of each point shows the clustering of behavior through the behavioral models by the SUBMODES system.
In this simulation the system learned 17 models.
Here a clear partition can be observed, where different behavioral models are active depending on the angular velocities $\omega_i$ and the turning velocity $\dot{ \alpha}$ (straight/left/right) of the point in behavioral space.
Note, that both $\omega_i$ and $\dot{ \alpha}$ were not directly available to the system, but instead the system used its internal predictions on changes of the sensory values to systematically structure the experienced behavior.

\begin{figure}
\begin{center}

\includegraphics[scale=0.3]{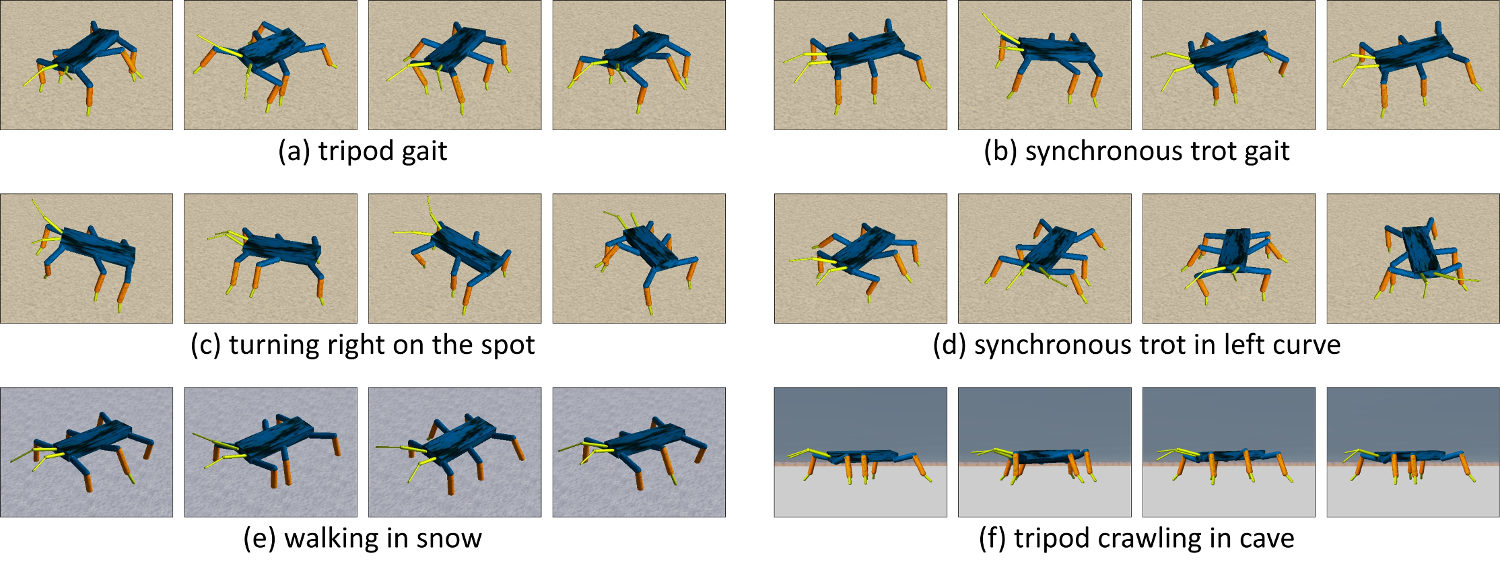}

\end{center}
\caption{Exemplary gaits discovered by the SUBMODES system for the Hexapod. Each gait was encoded by a single behavioral model $B_i$. (a)-(d) show gaits in an open field. (e) and (f) show gaits in different terrains (see section \ref{sectionObstacle}). In (e) snow slows down leg movements within it. In (f) a low ceiling limits the upward movement range of the legs. The inter-frame interval for the shown images is approximately 0.2 seconds.\label{hexapodGaits}}
\end{figure}

We tested the Hexapod robot in an open field without any obstacles.
When applied to the Hexapod the DEP-controller, with a particular inverse model $M$, generates different gaits with circular or oval forward movements of each leg.
The performed gaits vary in the strength of leg movements and the relationships of the phases between leg movements.
One of the emerging gaits for the Hexapod is the \textit{tripod gait}, as previously observed in \cite{dep1}.
The tripod gait, shown in Fig. \ref{hexapodGaits} (a), can be characterized as always having three legs on the ground and the ipsilateral front and back leg and the contralateral middle leg moving together
and in phase \cite{insectGaits}.
Moreover, a \textit{synchronous trot gait} could emerge, where two legs at opposing sides of the body move synchronously and hind and front leg movements are synchronized \cite{dep1}, as shown in Fig. \ref{hexapodGaits} (b).
Additionally, various hybrid forms of these gaits emerged, for example, front and middle legs moving as during the tripod gait and hind legs moving synchronized and in phase.
When activating the bias dynamics of the DEP-controller, the legs on one side of the body are offset either dorsally or ventrally alongside the rotational axes of the coxa joints.
This causes the legs on one side to rotate with a smaller amplitude, resulting in the robot crawling in a left or right curve, as shown in Fig. \ref{hexapodGaits} (c)-(d).

In 90 minutes exploring behavior for the Hexapod, the SUBMODES system learned on average 18 behavioral models ($\sigma = 2.9$) over 10 simulations.
Surprise is typically detected when the amplitude or phase-relation between the circular joint movements change, i.e., when the robot changes its gait, changes from crawling straight to crawling in a curve, or alters the overall velocity of the gait.
An example of changing from tripod gait to curved locomotion with the respective surprise-detection is shown in the lower row of Fig. \ref{surpriseFigure}.
Video 2 (\url{youtu.be/qeUpOqs9PCo}) shows more transitions in behavior for the Hexapod.

\subsection{Goal-directed locomotion}

In a second test we analyzed how the SUBMODES system can use its learned behavioral encodings for goal-directed planning and control.
We demonstrate this in a goal-reaching locomotion task.
In all experiments goals were small, circular areas.
Using an agent-centric frame of reference we define goals by means of a target orientation and velocity.
After either reaching the goal state or failing to reach it in time, the robot was reset and a new goal area was generated.
One simulation of this experiment consisted of 100 training episodes.
Each episode was composed of three different phases:
\begin{itemize}
\item \textit{Exploration phase:} During the exploration phase the system was allowed to discover and learn new types of behavior for five minutes of simulation time.
In this phase all motor commands were generated by the DEP-controller.

\item \textit{Training phase:} During the training phase the DEP-controller was deactivated and the motor commands were produced by the active behavioral models with the aim of reaching the given goal.
During the training phase the internal models of the system were updated.

In each training phase three goals were presented.

\item \textit{Testing phase:} The testing phase is equivalent to the training phase, except during testing no model updates occur.
This phase is included to measure the learning progress of the system over time.
Each testing phase consists of five randomly generated goal areas.
\end{itemize}
One training episode typically lasts between 12-17 minutes simulation time.

The Spherical robot (diameter $= 1$ unit) was tested in a large quadratic arena (size $ = 300 \times 300$ units) surrounded by walls.
Circular goal areas (radius $= 1$ unit) were randomly generated with a fixed distance around the center of the arena (distance $= 60$ units).
The Spherical robot was given a maximum of 140 seconds to reach a goal area before being reset.
Video 3 (\url{youtu.be/i0oovLnqF9A}) shows some exemplary runs.

\begin{figure*}
\begin{center}

\includegraphics[scale=0.37]{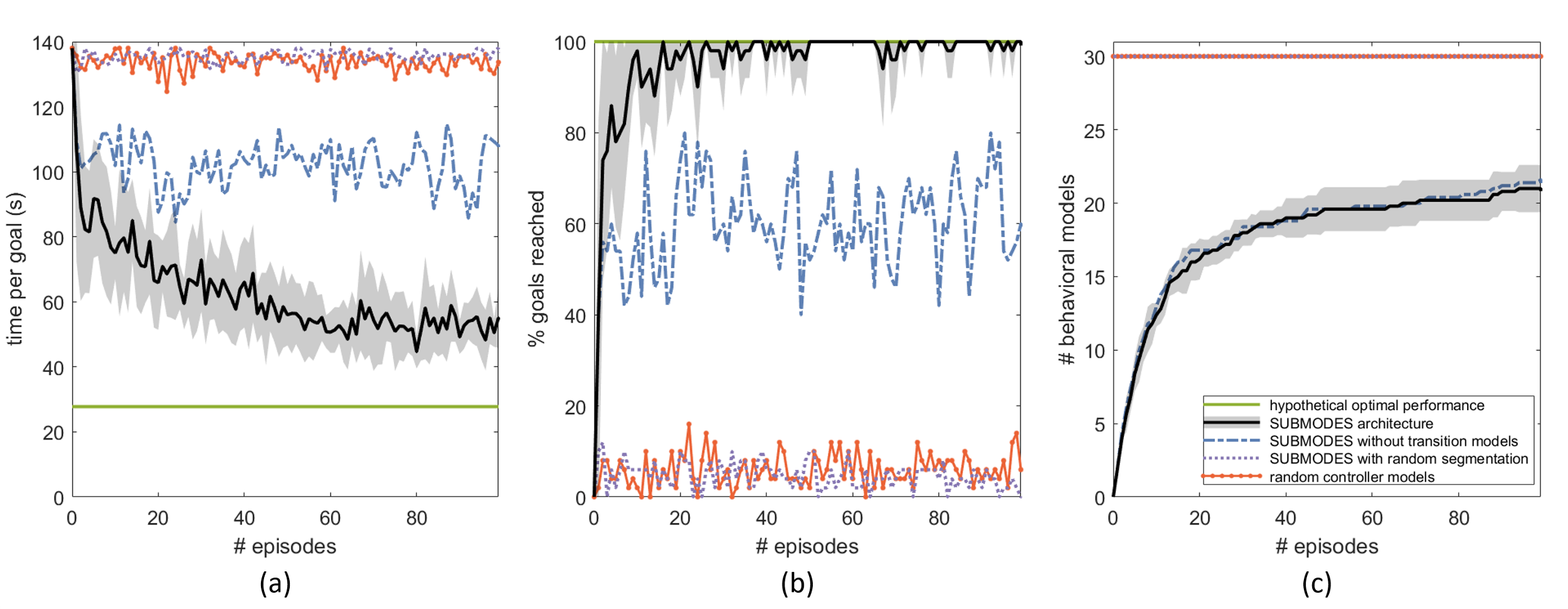}

\caption{Results for the goal-reaching task for the Spherical robot over the course of training episodes. (a) shows the average time spent per goal before the robot was reset. (b) shows the mean percentage of goal areas reached within the maximal time limit (140s). (c) shows the mean number of behavioral models discovered. The black line depicts the SUBMODES architecture with the shaded area showing the standard deviation. Other line styles and colors show different baselines (see text for further explanations).
\label{test2Spherical}}
\end{center}
\end{figure*}

Fig. \ref{test2Spherical} shows the results for the goal-reaching task for the Spherical robot, with the SUBMODES system shown in black.
Fig. \ref{test2Spherical} (a) shows the average time spent to reach the goal area.
Over the first 50 training episode the time required for goal-directed locomotion continuously decreases.
While in the first testing episodes the system required approximately 90 seconds per goal, during the last testing episodes it took less than 60 seconds.
As a reference, we include the hypothetical optimal performance of approximately 27 seconds that assumes no acceleration or turning is required and the robot can simply drive towards the goal with maximum speed.

Most of the behavioral models for the Spherical robot were discovered during the first 25 exploration phases, i.e., 125 minutes of exploring behavior.
The number of behavioral models increased only slightly afterwards (see Fig. \ref{test2Spherical} (c)).
Similarly, the percentage of goal areas reached within the maximal amount of time increased strongly over the first training episodes (see Fig. \ref{test2Spherical} (b)).
Already after the second training episode the SUBMODES system managed to reach over 70\% of the goal areas in time.
After 25 training episodes the system was able to reach more than 90\% of the goal areas.

We compare the performance of the SUBMODES system to different ablations of the system, also plotted in Fig. \ref{test2Spherical}.
To determine the effectiveness of self-organized exploration combined with surprise-based segmentation, we compare the system to behavioral control using \emph{random controllers}.
In this setting, the system does not explore its behavioral abilities but is instead equipped with 30 neural network controllers with fixed weights randomly generated following a uniform distribution ($\in [-1, 1]$).
The system can use these controller models for planning and goal-directed control.
Additionally, we compare the system to a \emph{random segmentation} baseline.
For this baseline the system is given 30 behavioral models $B_i$ and during exploration a randomly selected model is activated after each 5 seconds simulation time.
This baseline is used to determine the effect of surprise-based segmentation compared to random time-based segmentation.
Moreover, we tested the SUBMODES system \emph{without transition models}.
In this case, exploration and segmentation are applied normally, but no transition models are learned for transitions between behavioral primitives.
Thus, the system cannot know if a transition between two models is possible and cannot anticipate how a transition may affect its future sensory states.
This setting is included to test the effect of learning transition models for goal-directed planning.

As shown in Fig. \ref{test2Spherical}, the SUBMODES system clearly outperforms all of its ablations with respect to number of goals reached and time required per goal.
In the \emph{random controller models} setting, the system learns that some of the controllers can be used for locomotion, however, it finds no reliable way of changing direction.
As a result, using random controllers the robot only managed to reach goal areas if by chance it ended up with the right orientation towards the goal.
This is strongly reflected in the percentage of goal areas reached in time, which is on average below 20\% for all testing episodes.
Similar results can be observed for the random segmentation setting.
In this setting most of the learned models do not represent a consistent type of behavior.
Hence, the system managed to reach goal areas only by chance and, as a result, on average reaches less than  20\% of the goals during all episodes.
Without transition models the system did not only take more time to reach the goal areas, but also on average only reached approximately 60\% of the goal areas in time.
We assume, that without transition models the system makes errors in planning when predicting changes in behavior resulting in a worse performance.

The Hexapod robot (length $= 1$ unit) was tested in a large area without any obstacles.
Circular goal area (radius $= 1$ unit) were randomly generated  around the reset point of the robot with a fixed distance (distance $= 60$ units).
The Hexapod was reset if it did not reach a given goal area within 200 seconds simulation time.
Video 4 (\url{youtu.be/1h083TjLDK8}) shows some exemplary runs.

\begin{figure*}
\begin{center}
\includegraphics[scale=0.38]{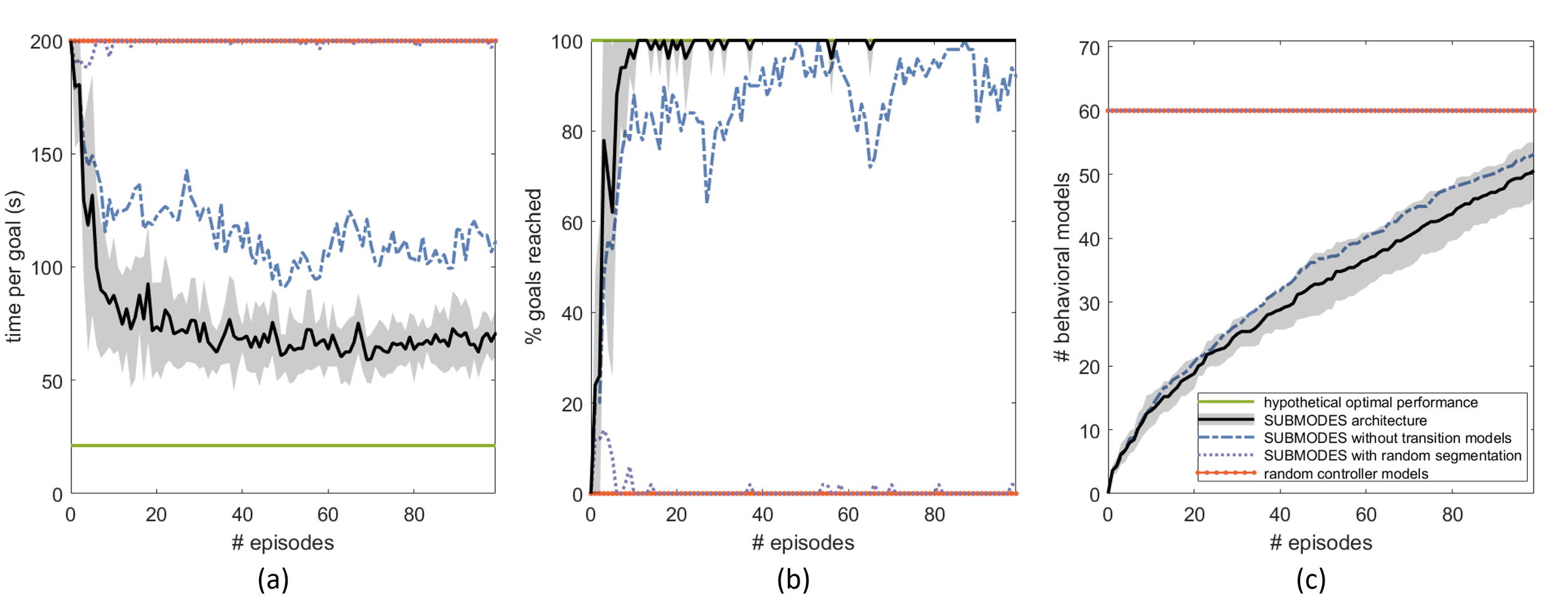}
\caption{Results for the goal-reaching task for the Hexapod over the course of the training episodes. (a) shows the average time spent per goal before the robot was reset. (b) shows the mean percentage of goal areas reached within the maximal time limit (200s). (c) shows the mean number of behavioral models discovered. The black line shows the performance of the SUBMODES architecture with the shaded area showing the standard deviation. Other line styles and colors show different baselines (see text for further explanations). \label{test2Hexapod}}
\end{center}
\end{figure*}

Fig. \ref{test2Hexapod} depicts the results of the goal-reaching task for the Hexapod robot when using the SUBMODES architecture (black line).
Already after the first training episode the system was able to reach 80\% of the goal areas within the maximal amount of time.
From the 10th training episode onward more than 90\% of the presented goal areas were reached in time.
The time required to reach the goal areas rapidly decreases over the first training episodes.
From the 60th episode onward all goals were successfully reached.
In the last testing episodes the system needed on average 70 seconds simulation time to reach the goal areas.
The hypothetical optimal performance of approximately 22 seconds simulation again assumes constant maximal speed directly to the goal, which cannot be reached.
The system continuously discovers new behavioral models over the course of the exploration phases.

As before, we compare the performance of the SUBMODES system to different ablations of the system, see Fig. \ref{test2Hexapod}.
When using \emph{random controller models}, the Hexapod never managed to reach a goal area.
While in some simulation we observed, that some random controllers could be used for changing the orientation of the robot, not once was a controller generated that could be used for locomotion.
Thus, using random controllers the Hexapod never managed to actually move to the goal areas.
When applying \emph{random segmentation} the robot reached approximately 10--15\% of the goal areas in time during the first two episodes, but only very rarely reached a goal area afterwards.
The cause for this could be that without the surprise-based segmentation one specific behavioral model does not correspond to a particular behavioral primitives, but instead each model is trained on various different types of behavior.
Even if by chance one model encodes a consistent behavioral primitive, it might get overwritten very quickly, resulting in a degeneration of performance.
As for the Spherical robot, the system \emph{without transition models} performs worse in the goal-reaching task in terms of time required to reach a goal area and number of goals reached in time.

\subsection{Terrain-dependent locomotion}
\label{sectionObstacle}
The previous tests showed that the SUBMODES system is able to identify self-explored behavioral primitives and learn models of these behavioral units and transitions thereof that can be applied for goal-directed locomotion.
In a third test we want to further examine if the system is also able to distinguish between different external events affecting the behavior of the robot.
For this purpose, we test it in an environment consisting of three different terrain types: a cave, an open field and a snow field.
The \textit{cave} has a low ceiling $1.1h$ above the ground with $h$ being the combined length of the Hexapod's tibia and tarsus.
Thus, the Hexapod is not able to fully lift its legs, when positioned in the cave.
However, the ceiling and the floor of the cave have a low friction, which allows the Hexapod to locomote forward using mostly forward-backward motions of its legs.
The second environment is an \textit{open field} without obstacles and a floor with normal friction (as in the previous experiments).
The third environment is a \textit{snow} environment.
In this environment a $0.4h$ tall snow layer is covering the ground.
All movements inside the snow layer are severely slowed down, by the factor $0.8$, caused by the high friction of the snow.

\begin{figure}
\begin{center}
\includegraphics[scale=0.2]{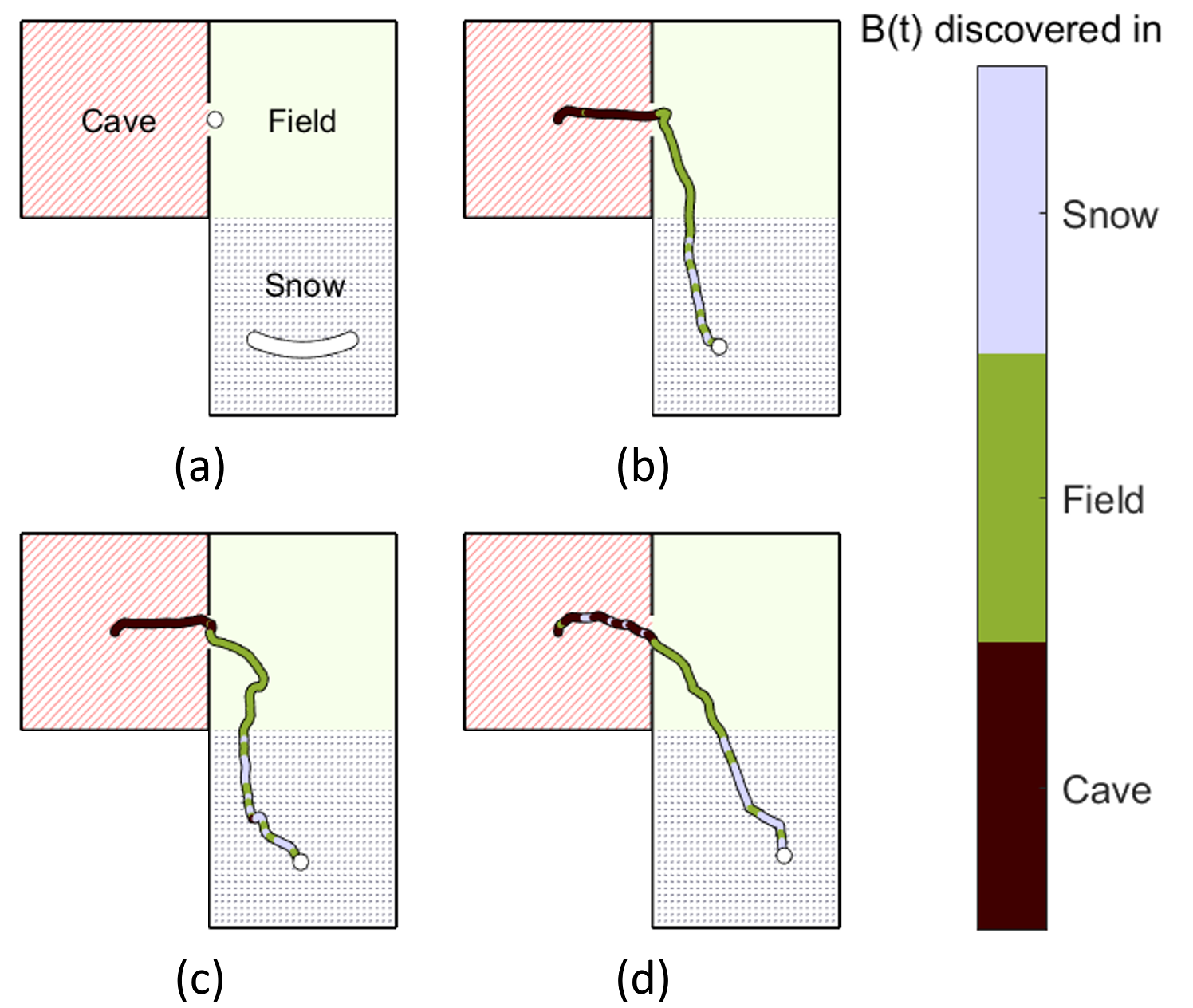}
\end{center}
\caption{Trajectories of the Hexapod for goal-directed locomotion in different terrain. (a) illustrates the obstacle course consisting of three different environments. Textures depict the type of environment and black lines represent walls. White areas show possible goal positions. The first goal is always positioned at the exit of the cave, the second goal is positioned inside the snow. (b)-(d) show exemplary trajectories from the last testing phases of different simulations. The color of the line denotes in which environment the used behavioral model was first discovered.  \label{obstacleCourse}}
\end{figure}

The SUBMODES system was given 60 minutes simulation time of behavioral exploration in each of the three environments.
Afterwards, the robot was placed in an obstacle course consisting of all three environment types (each with a size of $60 \times 60$ units), shown in Fig. \ref{obstacleCourse} (a), and had to use its previously learned models for goal-directed control.
The robot starts in the center of the cave facing the north wall.
The first goal is to crawl out of the cave through an opening at the right side of the cave.
After reaching the opening, a goal area was randomly positioned in the snow field and the task was to move over the open field and the snow layer to the goal position.
Fig. \ref{obstacleCourse} (a) shows the possible positions of the goal areas in white.
If the robot reached the goal area or did not reach it within an upper time limit (400 seconds simulation time), the robot was reset inside the cave.
Like in the previous tests, goal positions were defined with respect to the desired orientation $\alpha$ and velocity $v$ of the robot.
We tested the system for 100 training episodes, where each episode was composed of a training phase, during which one goal area was presented and the internal models were updated, and a testing phase, with five goal areas and without any model updates.

The SUBMODES system discovered new behavioral models for each of the three environments.
In the cave the system found different crawling motions, which allowed the Hexapod to move using only little upward movements of the legs.
One behavior that was discovered in the cave in every simulation is \textit{tripod crawling}.
During this behavior the legs are moved forward and backward as during the tripod gait, but only by slightly lifting its legs, as shown in Fig. \ref{hexapodGaits} (f).
In the snow environment, the system discovered interesting gaits for fast movement despite the high friction of the snow layer.
During most of the gaits discovered in snow, at least two legs are periodically lifted outside of the snow while the other legs move only little and their feet constantly stay within the snow, as for example shown in Fig. \ref{hexapodGaits} (e).
Some behavioral models were activated in more than one type of environment, but these behaviors mostly resemble standing still or performing little leg movement.
The system discovered on average $34$ behavioral models ($\sigma = 3.8$, $n = 10$) during the 180 minutes simulation time of exploration.
On average $9$ models were discovered in the cave ($\sigma = 2.3$), $15$ models in the open field ($\sigma = 3.5$), and $10$ models were discovered in the snow environment ($\sigma = 2.0$).

\begin{figure}
\begin{center}
\includegraphics[scale=0.4]{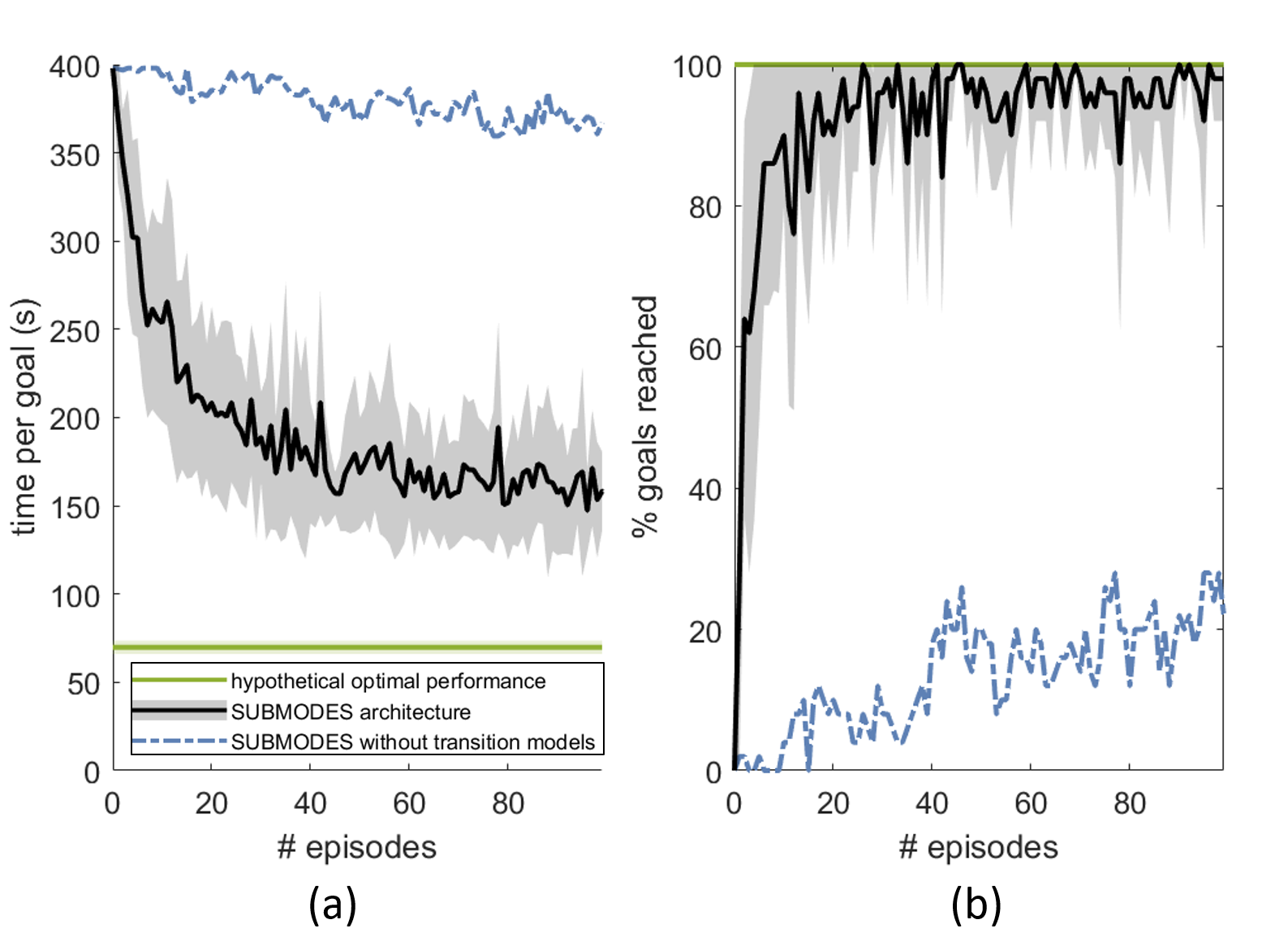}
\caption{Results for the terrain-dependent goal-reaching task for the Hexapod over the course of training epochs. (a) shows the average time spent per goal before the robot was reset. (b) shows the mean percentage of goal areas reached within the maximal time limit (400\,s). The solid black line depicts the SUBMODES system with the shaded area showing standard deviation; the dashed blue line shows the performance of the system without transition models; the solid green line shows an estimate of hypothetical optimal performance. \label{resultsObstacleCourse}}
\end{center}
\end{figure}

The results for goal-directed locomotion in the obstacle course are shown in Fig. \ref{resultsObstacleCourse}, with the black line depicting the SUBMODES architecture.
The time spent to reach a goal area and the percentage of goal areas reached by the system rapidly improves over the first couple of training episodes.
Already after seven training episodes the system was able to reach more than  80\% of the goal areas in time.
The percentage of goal areas reached in time further increased, such that the system reached more than 95\% of the goal areas during the last couple of episodes.
Furthermore, time spent to reach a goal area is approximately halved over the course of training.
Video 5 (\url{youtu.be/xhEmmm6VMg8}) shows one exemplary run of the Hexapod through the obstacle course.

In Fig. \ref{obstacleCourse} (b)-(d) some trajectories generated by the SUBMODES system for this task are illustrated.
The background pattern denotes the type of environment and the color of the lines show in which environment the active behavioral model was first discovered.
One can see that the system mostly applies behavioral models in one specific environment that were first discovered in this particular environment.
Hence, the system seems to distinguish between different types of behaviors based on the three different environments and learns which behaviors are applicable per environment.
Note, that the system does not receive direct information about its current environment.
The applicability of one behavioral primitive is determined purely by the prediction errors of the internal models and by learning the transition probabilities between different behavioral models.
The necessity of transition models for this task is clearly reflected in the performance of the ablated system without transition models (see Fig. \ref{resultsObstacleCourse}, blue line).
Without learning transition models, the system takes longer to improve its performance for goal-directed locomotion, and never reaches more than 30\% of the goal areas in time.

\section{Discussion and Future Work}

We have proposed a novel computational architecture, the SUBMODES architecture for surprise-based learning of modular, event-predictive behavioral primitives.
We showed through different simulations that this system is able to discover and detect a variety of behavioral primitives in highly complex, dynamic systems without the provision of any signal indicating the existence of a behavioral unit or the beginning or end of such a unit.
Instead, the system uncovered different behavioral primitives from a continuous self-explored sensorimotor stream in a self-supervised fashion purely based on the detection of surprise and principles of event-predictive cognition \cite{zacks, Butz:2016}.
This allowed our system to discretize the continuous stream of information experienced by an embodied agent online, while simultaneously learning models of the performed behavior and transitions in behavior.
In this way, the SUBMODES system was able to learn a repertoire of various behaviors for two complex robotic agents from scratch.

In this work, the behavioral capabilities were initially explored by means of self-organizing behavior, which was generated by the differential extrinsic plasticity (DEP) controller \cite{dep1}.
This controller was able to produce various complex, highly-coordinated behavioral patterns for the two robots with completely different body kinematics.
Without specifying a goal, various rolling motions for the Spherical robot and crawling behaviors for the Hexapod emerged, most notably the tripod gait also known from real insects \cite{dep1}.
While it has been shown before that DEP can discover and produce interesting types of behavior \cite{dep1, dep2}, the controller was, to the best of our knowledge, never used to bootstrap behavioral learning.
The SUBMODES system demonstrated that DEP is highly suitable for sensorimotor exploration in a self-supervised learning architecture.
However, the SUBMODES architecture does not rely on this particular controller.
Other forms of behavioral exploration or learning by demonstration could in principle be applied as well, including predictive information maximization \cite{PIMAX}, intrinsically motivated goal exploration processes \cite{forestier2017intrinsically}, or human demonstration.

Traditionally, learning behavioral primitives was investigated by learning only one primitive in isolation or by providing either explicit labels of the ongoing primitives or labels signaling transitions between primitives \cite{schaalMovementPrimitives,Ijspeert:2013,Nguyen-Tuong:2011, Worgotter:2013, naturalActorCritic, Calinon:2009,Kober:2011, Sigaud:2011}.
Our system segments behavioral primitives without any supervised information or explicit labels.
Classical approaches for self-supervised, online segmentation of behavior were applied in much simpler toy-scenarios or in sensory spaces with a lower complexity \cite{wolpert2, surprise, Simsek:2004,Simsek:2009}.
Related systems for learning predictive behavioral encodings for more complex robotic systems learn based on replays of manually-demonstrated primitive motions, e.g., \cite{tani2003, tani2008}, which simplifies the segmentation problem because the trajectories during training have smaller variations such that transitions are more apparent.
We have shown that our surprise-based segmentation mechanism works well for high-dimensional, noisy, self-generated streams of sensorimotor information.

Besides the segmentation and learning abilities, we showed that the SUBMODES system can use its learned behavioral representations progressively more effectively for solving various goal-reaching tasks.
The improvement in performance over time is accomplished by three main mechanisms:
(1.) Over time, the system discovers new types of behavior, which may be more effective for the tested tasks.
(2.) The system continues to improve the accuracy of the available behavioral models, enabling the more accurate anticipation of sensory consequences for each associated behavior.
(3.) The system improves its predictive models about behavioral transitions, learning when transitions between different types of behavior can be applied and how a specific transition affects the sensory state.

The learning of modular behavioral models paired with sensorimotor exploration, allows the SUBMODES system to rapidly acquire models suitable for goal-directed control.
This cannot be achieved by model-free approaches of behavioral control, such as model-free RL methods.
For example, when applying Soft Actor-Critic to the Mujoco Ant-v1---a four-legged robot similar to the Hexapod but with fewer degrees of freedom---more than 1M update steps were needed to achieve reasonable forward locomotion alone \cite{SAC}.
In comparison, the SUBMODES system applied to the Hexapod managed to learn at least one good model for locomotion already during the first exploration phase, which takes less than 15k update steps.
Hence, our system seems to be roughly two orders of magnitude faster than a state-of-the-art deep RL approach when applied to similar robots.
On top of that, our system learns additional models for locomotion and turning and is able to use the learned models for inducing flexible goal-reaching of target areas within a restricted time interval.
A success rate of more than 90\% was reached in less than 400k update steps.

While the system manages to improve its capabilities to perform goal-directed control both in terms of the number of goal states reached and the time required to reach these goals, the system currently does not quite achieve optimal performance for the examined tasks.
However, note that the learned representations were not optimized for any of the tested objectives.
Instead, the system learned general, abstract representations of behavior that can in principle be applied in various tasks.

If one wishes to further optimize the performance of SUBMODES with respect to a specific task, there are various methods that could be applied in addition to the already involved processes:
Seeing that the learned models are differentiable, applying goal-directed active inference is possible \cite{Otte:2017, Friston:2015}, adjusting the motor command of each behavioral model depending on the desired sensory outcome on the fly.
Furthermore, if a criterion for successful performance in a specific task is known, for example achieving high velocity in a locomotion task, the models could be optimized to further achieve this criterion by means of model-free RL \cite{SuttonBarto2018:RLIntro} and policy gradient approaches \cite{naturalActorCritic}.

SUBMODES modularizes the experienced behavior by encoding behavioral primitives through discrete, individual models.
While this modularization protects the system from catastrophic forgetting \cite{catastrophic1}, the time required to learn different behaviors could be further improved by sharing information among models.
Hence, for future work we want to apply the principles employed here to a more general forward architecture, akin to the network architectures in \cite{Tani:2017}, and explore how behavioral representations can be modularized by selectively activating sub-components within the same network structure, as for example demonstrated by the REPRISE architecture \cite{Butz:2018, butz2018learning}.

Besides further behavioral optimization and a less strict modularization, we intend to explore the applicability of SUBMODES to more complex tasks.
One challenge in this respect is higher task complexity, where multiple intermediate goals need to be accomplished to reach a desired final goal state---such as when the hand first needs to move to a bottle before moving the hand to the mouth in order to drink out of the bottle.
We expect that such tasks require non-greedy, conceptual planning mechanisms that unfold on deeper, conceptualized levels of abstraction.
Finally, we intend to tackle the visual sensory challenge and apply SUBMODES to real robots, where precise location, state, and motion information are not available but need to be inferred from the given sensory information indirectly.

\bibliographystyle{unsrt}

\bibliography{bibFileBehavioralPrimitives} 

\clearpage

\appendix
\section{Behavioral exploration using DEP} \label{sectionDEP}
\begin{figure}
\begin{center}
\includegraphics[scale=0.13]{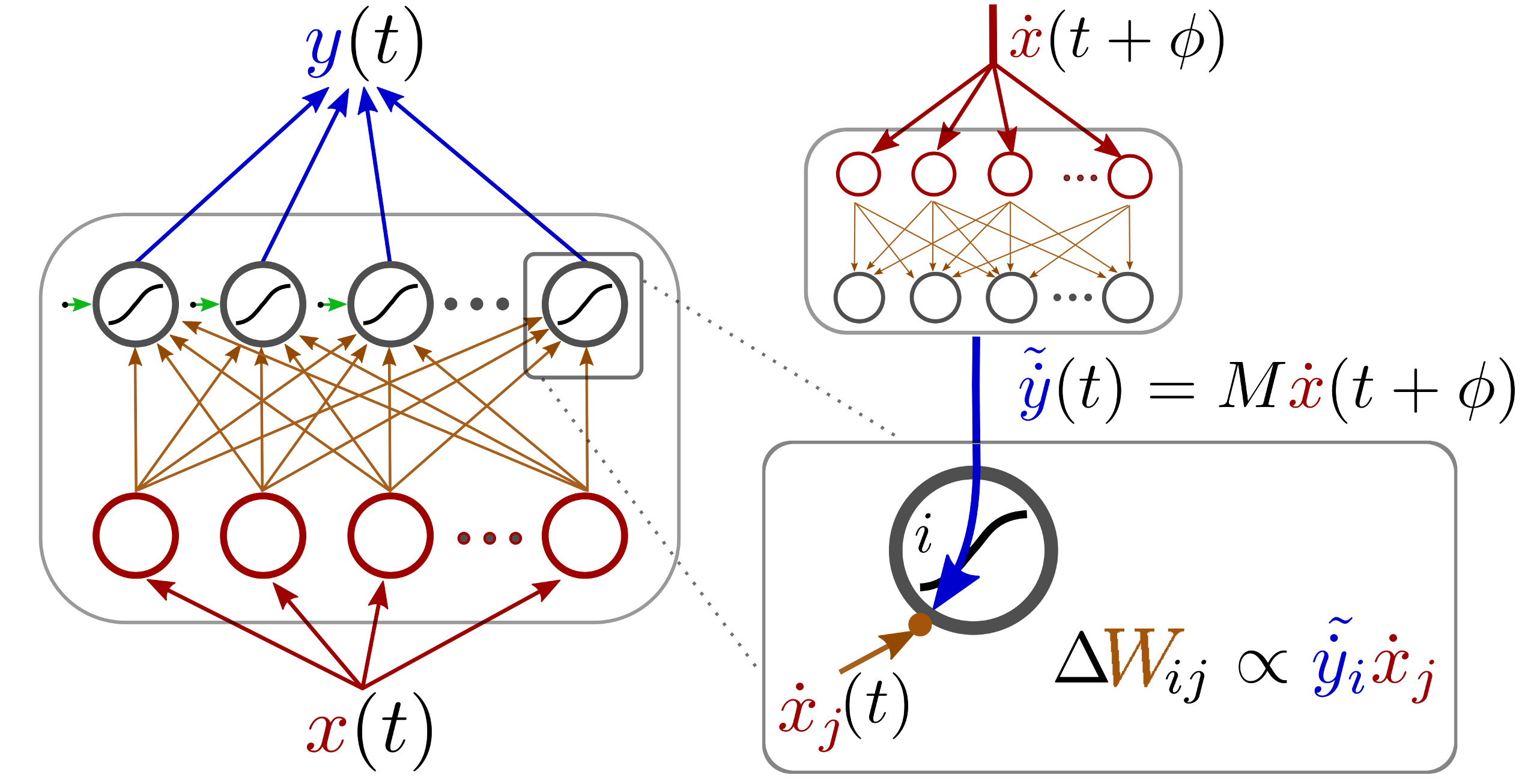}
\end{center}

\caption{Network architecture of the DEP-controller (adapted from \cite{dep2}). The left side illustrates the neural network controller generating motor commands $y(t)$ based on the proprioceptive sensory input $x(t)$. The right side shows the DEP lear\-ning rule, multiplying the derivative of a sensor value $\dot{x}(t)$ with the inferred motor changes ${\tilde{\dot{y}}}(t)$, generated by the inverse model $M$ from some future input's derivative $\dot{x}(t + \phi)$. \label{dep}}
\end{figure}

For the parametric setup of the DEP-controller we follow \cite{dep1}.
The complete controller architecture is illustrated in Fig. \ref{dep}.
The DEP-controller receives an $n$-dimensional sensory input $x(t)$ and  generates an $m$-dimensional motor command $y(t)$ at every discrete time step $t$.
We assume that the system has a basic understanding of the causal relationship between motor actions and proprioceptive sensor values \cite{dep1}.
This `understanding' is imprinted into an inverse model $M$, which relates sensory values $x(t + \phi)$ back to motor commands $y(t)$ with a certain time lag $\phi$.
When focusing on changes in sensory values and motor values, we get
\begin{equation}
\tilde{\dot{y}}(t) = M\dot{x}(t + \phi), \label{eqM}
\end{equation}
where $M$ is the inverse model, simplified as a linear model in the form of a $m \times n$ matrix, and the time lag $\phi = 1$.

The controller weights are then updated using the differential extrinsic plasticity rule (DEP):
\begin{equation}
\Delta W_{ij} = \epsilon_W \hspace*{1pt} (\tilde{\dot{y}}_i(t) \dot{x}_j(t) - W_{ij}), \label{eqCupdate}
\end{equation}
where $\epsilon_W = 0.1$ is a learning rate and $-W_{ij}$ is a damping term.
Since $\tilde{\dot{y}}(t)$ is a linear transformation
of $\dot{x}(t + \phi)$, the synaptic weights of the controller change based on correlations between changes in sensor values $\dot{x}$ with a time lag $\phi$. Thereby, the inverse model $M$ states how correlations between $\dot{x}_i(t + \phi)$ and $\dot{x}_j(t)$ impact the weights $W$.

As in \cite{dep1} we use an appropriate normalization  of the controller weights $W$.
There are two options to perform weight normalization: \emph{global normalization} and \emph{individual normalization}.
For global normalization the entire weight matrix is normalized:
\begin{equation}
W \leftarrow \kappa \frac{W}{||W|| + p}
\end{equation}
with $\kappa$ an empirical gain factor and a regularization term $p = 10^{-12}$ that becomes effective near the singularity ($||W||=0$).
In individual normalization each motor neuron is normalized individually, with
\begin{equation}
W_{ij} \leftarrow \kappa \frac{W_{ij}}{||W_i|| + p},
\end{equation}
where $||W_i||$ is the norm of the $i$th row of $W$, consisting of all weights that connect to
the motor neuron $i$.
The type of normalization applied has a strong effect on the resulting behavior:
While individual normalization leads to behaviors that involve all motors, global normalization restricts the overall activity to a subset of motors.
For the Spherical robot we apply global normalization, which results in the behavior in which two internal masses are constantly moved while the third mass is stationary.
For the Hexapod robot we apply in\-di\-vi\-dual normalization, resulting in all joints being involved for locomotion.
The gain factor $\kappa$ regulates the overall feedback strength of the sensorimotor loop, which we set to
 $\kappa = 1.5$ for the Spherical robot and to $\kappa = 2.2$ for the Hexapod.

The controller additionally uses a bias dynamics that we added to the original DEP-controller.
The bias dynamics changes the values of one bias neuron every $\tau_h$ time steps.
The bias ought to be altered is chosen as the bias neuron connecting to the motor neuron $i$ that has had the fewest changes in the controller weights $W_{ i}$ connected to the neuron $i$.
Based on this heuristics we introduce activity to motor neurons that did not change their activity much in the recent past.
When the bias neuron is activated its activity is randomly set to either $h_i = 1.5$ or $h = -1.5$.
After $\tau_h$ time steps all bias neurons are deactivated again and this process is repeated.

For the Spherical robot we set $\tau_h = 5000$ (100 seconds).
Since for the Hexapod robot the behavior demonstrated by the DEP-controller changes more often naturally, we chose a larger time horizon of $\tau_h = 10000$ (200 seconds).
The DEP-controller applied to the Spherical robot uses three bias neurons, one for each motor neuron.
For the Hexapod we use four bias neurons.
Two bias neurons are connected to the forward-backward coxa joints of either the right legs or the left legs and two bias neurons are connected to the upward-downward coxa joints of either the right or left legs.
With this wiring, the activation of one bias neuron can offset the coxa joint positions of the legs on one body side to four different directions (upward, downward, forward, backward).

The inverse model $M$ of the DEP-controller, states how sensory changes relate back to changes in the motor commands of the system, as defined by Equation \ref{eqM}.
If the DEP-controller uses only proprioceptive sensory information as an input and motor commands of the same joints as an output, we can set $M = \mathbb{I}$ to the identity matrix $\mathbb{I}$.
This design corresponds to the idea that changes in the proprioception of joint $i$ are caused by changes in the motor command of joint $i$.
This setting can be considered the standard case of applying the DEP-controller, which we also use for the Spherical robot.

\begin{figure}
\begin{center}
\includegraphics[scale=0.25]{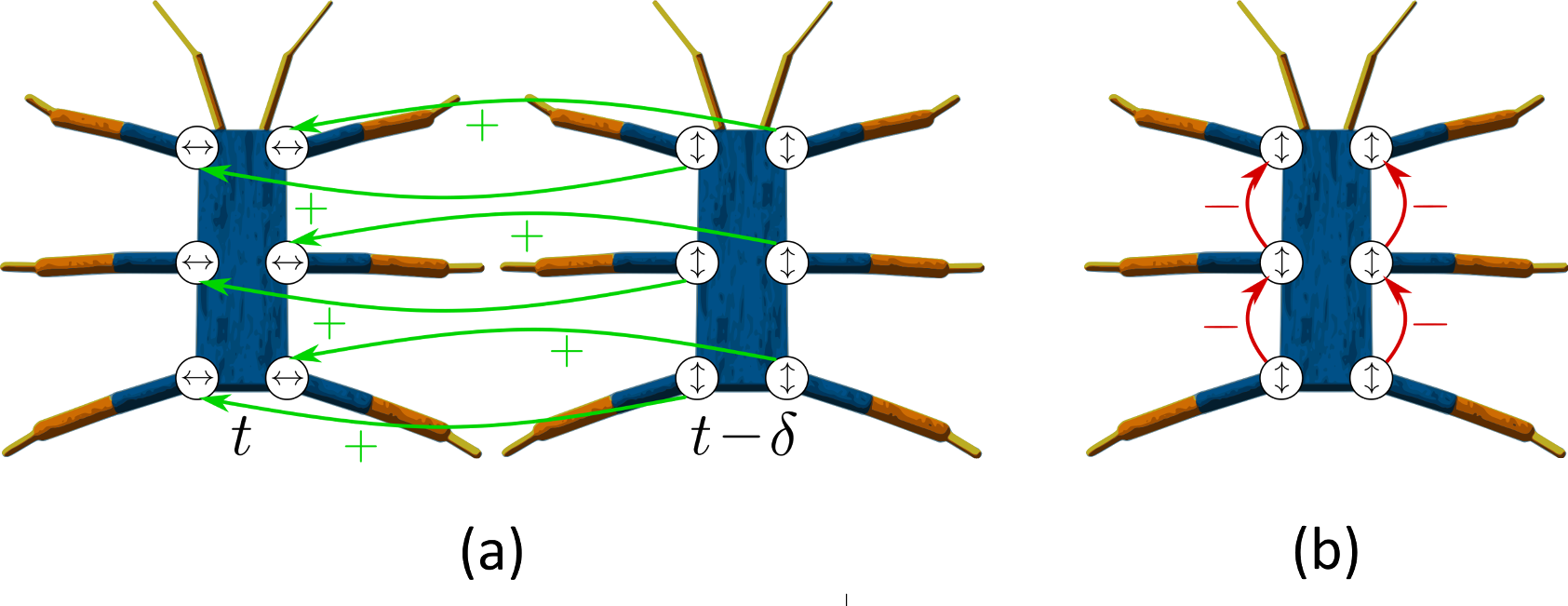}
\end{center}
\caption{Prestructuring of the inverse model $M$ of the DEP-controller when using the Hexapod. $\leftrightarrow$ depicts the up-down dimension of the coxa joint, $\updownarrow$ depicts the forward-backward dimension. An arrow from joint $i$ to $j$ describes the entry $M_{ij}$ of the inverse model matrix. $+$ -arrows represent a positive connection ($M_{ij} = 1$), $-$ -arrows represent a negative connection ($M_{ij} = -1$). \label{inverseModelHexapod}}
\end{figure}

However, the inverse model $M$ can also be prestructured, by adding connections between joints within the inverse model $M$ where correlations or anticorrelations of the joint velocities are desired.
The underlying idea is, that we can add connections for joints $i$ and $j$ to increase either positive
correlations ($M_{ij} > 0$) or negative correlations ($M_{ij} < 0$) between their velocities over time.
We apply this form of \textit{guided self-organization of behavior} \cite{guidedSelfOrganization} when using the Hexapod, as it was previously done by \cite{dep1}.
For the Hexapod, the inverse model $M$ assumes a positive correlation between changes of joint angles and changes in motor commands for the same joint, i.e., $M_{ii} = 1$ for a joint $i$.
Furthermore, the time-delayed sensor for forward-backward angles is positively linked to the downward-upward angle of the same coxa joint (see Fig. \ref{inverseModelHexapod} (a)). This connection facilitates circular leg motions over time \cite{dep1}, e.g., once the leg moves forward it is desired that the leg moves downward some time later.
To further facilitate locomotion we additionally want to obtain antiphasic forward-backward motions of subsequent legs on the same side.
For this purpose, negative links are included in $M$ between the forward-backward sensors and motors of subsequent legs
of the same side (see Fig. \ref{inverseModelHexapod} (b)).

\section{Learning transitions in behavior} \label{sectionTransitions}

To enable the accurate prediction of a sensorimotor time series consisting of a variety of different behaviors, it is important to not only consider how the sensorimotor information unfolds during each stable behavioral mode, but to also model the transitions between two subsequent behavioral primitives.
For this purpose, the SUBMODES system incorporates a set of transition models $\mathcal{T}$.
If a transition from model $B_i$ to model $B_j$ occurs the transition model $T_{i\rightarrow j} \in \mathcal{T}$ is updated.
Each transition model consists of three subcomponents: $P_{i \rightarrow j}$, $\bar{t}_{i \rightarrow j}$, and $F_{i \rightarrow j}$.

Some transitions require a specific context to occur, e.g., a transition from \lq walking\rq \space to \lq swimming\rq \space can only occur if the agent is standing in shallow water.
To model the critical conditions for a transition in behavior,  $T_{i\rightarrow j}$ contains a transition probability network $P_{i\rightarrow j}$.
This network aims to predict the probability of a successful transition from $B_i$ to $B_j$ given the current sensory state $x(t)$.
$P_{i \rightarrow j}$ is a single layered feed-forward neural network mapping a sensory state $x(t)$ to a probability $\in [0, 1]$.
If a transition was initiated at time step $t$, then $P_{i \rightarrow j}$ receives $x(t)$ as an input to train the network.
If after the transition the system activated model $B_j$, then $P_{i \rightarrow j}$ is trained on the deviation of its prediction from the target probability $1$.
If the system planned to reach $B_j$ when initiating the transition, but ended up using a different model, $P_{i \rightarrow j}$ is updated using the target probability $0$.
Thus, the network estimates the probability of being able to switch from $B_i$ to $B_j$ given the current sensory state.

Transitions in behavior may take different times to be completed, since every transition in behavior is preceded by a searching period.
Hence, $T_{i \rightarrow j}$ contains the component  $\bar{t}_{i \rightarrow j}$, an estimation of the time required to perform a transition from $B_i$ to $B_j$.
Currently $\bar{t}_{i\rightarrow j}$ is computed as the mean time steps which passed between the initiation of a transition from model $B_i$ and successively activating model $B_j$.

Transitions in behavior can also entail a strong sudden sensory change.
For example a transition from \lq running\rq \space to \lq standing still\rq \space typically results in a strong decrease in velocity.
To predict the sensory changes occurring during a transition between models an additional single-layered feed-forward neural network $F_{i \rightarrow j}$ is trained.
$F_{i \rightarrow j}$ learns a mapping from a sensory state $x$ to a change in sensory states $\Delta x$.
When a transition from model $B_i$ is initiated at time $t$ and model $B_j$ is activated at time step $t + t_{i\rightarrow j}$ then $F_{i\rightarrow j}$ is trained on the input $x(t)$ and the nominal output $x(t + t_{i\rightarrow j}) - x(t)$.
Hence, $F_{i\rightarrow j}$ predicts how the sensory state will change from the onset of a transition until the transition is finished.

\begin{figure*}
\begin{center}
\includegraphics[scale=0.175]{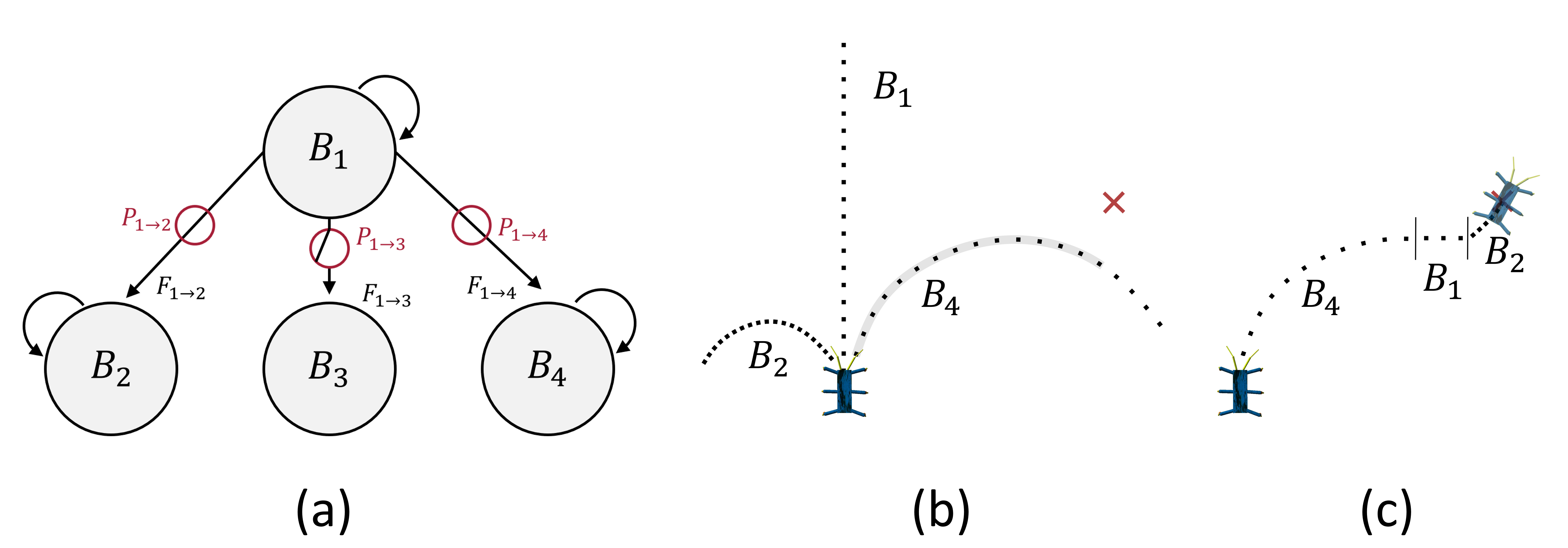}

\end{center}
\caption{Illustration of the representations of behavior learned by the SUBMODES architecture and their use for goal-directed planning. (a) The learned representations form a directed graph with the behavioral models $B_i$ as nodes. Each edge represents one step of sensory prediction, either by staying in the same model, or by transitioning to a new behavior.
A transition to behavior $B_j$ from current behavior $B_i$ is considered in a stochastic fashion according to the probability $P_{i \rightarrow j}$ given the current sensory state $x(t)$.
In this example $B_1$ is active and $B_2$ and $B_4$ can be reached. (b) shows how the prediction can be used for greedy planning. $\times$ marks a goal state and the dotted lines show the predicted trajectory when using an associated behavioral model $B_i$. In this example the system chooses $B_4$ since a part of the predicted trajectory, marked by a grey background, has the lowest mean distance to the goal state. (c) shows how replanning allows the system to concatenate different behavioral primitives for accurate goal-directed control. \label{plan}}
\end{figure*}

Overall, one transition model $T_{i\rightarrow j}$ can be used to estimate (1.) where in sensory space such a transition is applicable, (2.) how long the transition in behavior will take until the next model is active, and (3.) how the sensory state will change over the course of the transition, by means of $P_{i \rightarrow j}$, $\bar{t}_{i \rightarrow j}$, and $F_{i \rightarrow j}$, respectively.
This results in a directed graph representation of behavioral primitives, as illustrated in Fig. \ref{plan} (a).
Each node of the graph represents a stable behavioral mode with uniformly unfolding sensorimotor dynamics, encoded by a single behavioral model $B_i$.
The edges between two nodes are transitions in behavior, represented by a transition model $F_{i \rightarrow j}$.
The availability of an edge given the current sensory state, is encoded by the transition probability model $P_{i \rightarrow j}$.
This graph representation of behavior and transitions in behavior is crucial to allow hierarchical, goal-directed planning of behavior.

\section{Goal-directed planning} \label{sectionPlanning}

When switching into goal-directed control, the explorative controller is deactivated and behavioral models and model transitions are invoked purposefully for minimizing the difference between anticipated and desired perceptions.
This process of greedy planning is schematically illustrated in Fig. \ref{plan} (b).

During goal-directed control at time $t$ the system considers which subset of behaviors $\mathcal{B}(t) \subseteq \mathcal{B}$ is applicable given the current sensory state $x(t)$ and the currently active model $B(t) = B_i$.
Whether a behavior $B_j$ is an element of $\mathcal{B}(t)$ is determined stochastically using the transition probability network $P_{i \rightarrow j}$.
The system determines the probability $P(B_j \in \mathcal{B}(t) | B_i, x(t))$ of $B_j$ being an element of $\mathcal{B}(t)$ as
\begin{equation}
P(B_j \in \mathcal{B}(t) | B_i, x(t)) = P_{i \rightarrow j} ( x(t)), \label{eqPlan1}
\end{equation}
with $B_i$ the active model and $x(t)$ the current sensory state.

As a next step the system predicts how its sensory state will change when transitioning from the current behavior $B_i$ to a new behavior $B_j \in \mathcal{B}(t)$.
The sensory state $x'(t + \bar{t}_{i \rightarrow j}|B_i \rightarrow B_j)$, describing the sensory state after a transition to $B_j$ from the active model $B_i$, is determined as
\begin{equation}
x'(t + \bar{t}_{i \rightarrow j} |B_i \rightarrow B_j) = x(t) + F_{i \rightarrow j}(x(t)), \label{eqPlan2}
\end{equation}
with $\bar{t}_{i \rightarrow j}$ the estimated time required for the transition and $F_{i \rightarrow j}$ the transition network predicting the sensory change during a transition from $B_i$ to $B_j$.

Then, the system predicts how the sensory information will evolve over a planning horizon $\tau_{p}$ when staying in $B_j$.
The succeeding sensory states $x'(t+u |B_i \rightarrow B_j)$ are computed iteratively via
\begin{equation}
\begin{aligned}
x'(t + u + 1 | B_i \rightarrow B_j) \leftarrow  & x'(t + u |B_i \rightarrow B_j) \\
& + B_j(x'(t +u |B_i \rightarrow B_j)), \label{eqPlan3}
\end{aligned}
\end{equation}
starting with $u = \bar{t}_{i \rightarrow j}$ until $u = \tau_{p}$.

Given a goal state $x^G(t)$ the distance of a predicted sensory state $x'(t +u)$ with respect to the goal can be computed as
\begin{equation}
D^G(x'(t +u)) = |x^G(t) - x'(t +u)|_\mathcal{M}, \label{eqLoss}
\end{equation}
for some metric $\mathcal{M}$.
In the current experiments $\mathcal{M}$ was chosen as the squared distance between task-relevant sensory information of $x^G(t)$ and $x'(t +u)$.
In our examples, the task-relevant coordinates are the orientation $\alpha$ and the velocity $v$.

The next behavioral model $B(t+1)$ is chosen as
\begin{equation}
\begin{aligned}
B(t+1) \leftarrow & \argmin_{B_j \in \mathcal{B}(t)} \min_{\bar{t}_{i \rightarrow j} < \tau < \tau_{p}} \frac{1}{(\tau - \bar{t}_{i \rightarrow j})} \\
& \sum_{u =  \bar{t}_{i \rightarrow j}}^{\tau} D^G(x'(t + u |B(t) \rightarrow B_j)). \label{eqPlan}
\end{aligned}
\end{equation}
Hence, the next model $B(t+1)$ is determined, as the applicable behavior that predicts the sensory time series with the lowest mean distance to the goal.
The predicted time series has a maximal length of $\tau_{p}$ to ensure an upper limit on the computational complexity, which we set to $\tau_{p} = 500$.

After activating the next model $B(t+1)$, the system initiates a searching period to determine whether the transition to this model was successful, as described in section \ref{Submodes}.
The transition model $T_{i \rightarrow j}$ is then updated, depending on the success of the initiated transition.
As soon as the system is certain about which behavioral model is currently active, i.e., after the searching period, it is allowed to replan.
In this way, the system can serially concatenate single behavioral primitives to form a chain of more complex behavior that allows the system to accurately reach a given goal state, as illustrated in Fig. \ref{plan} (c).

\section{Parametric setup} \label{sectionParameters}

All neural network models of our system, i.e., $B_i, F_{i\rightarrow j}, P_{i \rightarrow j}$, are single-layered neural networks mapping directly from sensory input space $\mathcal{X}$ to their respective output spaces ($B_i: \mathcal{X} \times \mathcal{Y}, F_{i \rightarrow j}: \mathcal{X},P_{i \rightarrow j}: [0,1]$).
For networks predicting sensory changes or motor commands, i.e., $B_i$ and $F_{i \rightarrow j}$, output neurons use a $\tanh$-activation function and a squared error loss is used for back propagation.
To enforce sparsity in the network weights, a $L_1$ weight regularization term is added to the loss-functions \cite{L1Regularization}, with the regularization constant $\lambda = 0.005$.
For networks predicting probabilities, i.e., $P_{i \rightarrow j}$, output neurons use a sigmoid activation function and perform back-propagation based on a balanced cross-entropy loss \cite{balancedCrossEntropy}.
The different types of networks use different learning rates ($B_i: \epsilon_B = 0.005, F_{i \rightarrow j}: \epsilon_F = 0.01, P_{i \rightarrow j}: \epsilon_P =0.05$).
To enable fast learning while avoiding local overfitting, each network is equipped with a \textit{replay buffer} with a large capacity (capacity = 10000) that stores a new input-output pair in each training step.
During each network update $s$ additional samples are randomly drawn from the buffer and the neural network models are additionally trained on the drawn samples.
For the Spherical robot $s=2$ samples are additionally drawn during each network update.
Seeing that the behaviors change faster for the Hexapod, we used a larger sampling rate of $s=25$ in that scenario.

The error models $E_i \in \mathcal{E}$ are estimated as a normal distribution.
The normal distributions are initialized with $\bar{e}_i(0) \leftarrow 0.05$ and $\bar{\sigma}_i(0) \leftarrow 0$.
To allow each error model to quickly keep track of the prediction accuracy of its respective behavioral model, $\bar{e}_i$ and $\bar{\sigma}_i$ are updated as an exponential moving average and variance, with a timescale of 1000 steps ($\epsilon_E = 0.001$).

To enable the detection of surprise we compute the prediction error $e(t)$ as a simple moving average of the sensory prediction error over a short time interval (25 time steps or 0.5 seconds).
Comparing the prediction error $e(t)$ to the active error model $E_i$, allows the system to detect surprise (as defined in equation \ref{surprise}).
The surprise threshold $\theta$ determines the confidence threshold above which an error is considered \lq surprising\rq.
Seeing that we face highly noisy scenarios in our experiments, we chose a small threshold $\theta = 2$ to achieve a fine-grained segmentation.
However, depending on the general predictability of the scenario and the desired level of abstraction, a larger $\theta$ can be applied as well \cite{gumbsch2}.

Upon detecting surprise, the system enters a searching period with a duration $\in [\tau_{s,min}, \tau_{s,max}]$  to determine the next behavioral model. In our simulations the searching period takes at least $\tau_{s,min} = 50$ time steps (1 second), to have a sufficient number of data points for comparing the predictions of all models.
The searching period takes maximally $ \tau_{s,max} = 700$ time steps (14 seconds) before a new model is created.
Seeing that the DEP-controller maintains one type of behavior for a relatively long time (typically longer than a minute), we can use such a long searching period.
In this way, small irregularities in behavior, such as the Hexapod stumbling, are ignored instead of resulting in the generation of a new behavioral model.
However, for other exploration mechanisms with faster changes in behavior a shorter searching period is recommended.

\section{Processing sensory changes}
To allow the surprise-based segmentation to take all sensory dimensions into account equally, it is necessary that every sensory dimension $x_i$ changes at a similar rate. This can be achieved in two ways: (1.) choosing an appropriate time frame for determining the change in sensory information $\Delta x$, (2.) scaling each dimension $i$ of $\Delta x$ by a constant factor $c_i$, such that all $\Delta x_i$ are within the same interval.

For the Spherical robot $\Delta x$ is computed as the change of sensory information over one time step (i.e., $\Delta x(t+1) = x(t+1) - x(t)$). For the Hexapod $\Delta x$ is computed as the mean change over 10 time steps.
By computing $\Delta x$ in this way, changes in proprioception are typically within the same interval ($\Delta x_i \in [-0.1, 0.1]$).
To assure that other changes in sensory information are within this interval as well, $\Delta \sin(\alpha)$, $\Delta \cos(\alpha)$, and $\Delta v$ are multiplied with a constant factor $c$ ($c = 10$ for the Spherical robot, $c = 15$ for the Hexapod).

\section{Pseudocode} \label{sectionPseudocode}
In this section we provide pseudocode for the SUBMODES algorithm.
Algorithm \ref{algorithmMain} describes the main loop of the system.
Algorithms \ref{algorithmCreate} -- \ref{algorithmPlan} are separated from the main algorithm to improve readability.
Algorithm \ref{algorithmMain} receives the surprise threshold $\theta$, the minimal and maximal duration of the searching interval $[\tau_{s,min}, \tau_{s,max}]$, and the planning horizon $\tau_p$ as input parameters.

 \begin{algorithm}
 \caption{SUBMODES: main algorithm}\label{algorithmMain}
 \begin{algorithmic}[1]
 \Procedure{SUBMODES}{$\theta$, $\tau_{s,min}$, $\tau_{s,max}$, $\tau_p$}
 \State $t \gets 0$
 \State $B_0 \gets$ \textsc{create\_ new\_ model}$()$
 \State $\mathcal{B} \gets \{B_0 \}$, $\mathcal{E} \gets \{E_0 \}$, $\mathcal{T} \gets \{ \}$
 \State $B(t) \gets B_0$, $E(t) \gets E_0$
 \State initialize $DEP$
 \State $x(t) \gets$ sense current sensory state
 \State $y(t) \gets$ action from $DEP$ given $x(t)$
 \State $p(t) \gets$ prediction from $B(t)$ given $x(t)$
 \State $e(t) \gets 0$ \textcolor{black!50}{\Comment{prediction error}}
 \State $\textit{exploration} \gets$ true \textcolor{black!50}{\Comment{exploration vs. planning?}}
 \State $\textit{searching} \gets$ false \textcolor{black!50}{\Comment{in searching phase?}}
 \State $t_s \gets 0$ \textcolor{black!50}{\Comment{time spent searching}}
 \State execute action $y(t)$
 \While {simulation is running}
 \State \textcolor{black!50}{\Comment{1. surprise detection and model updates}}
 \State $t \gets t + 1$
 \State $x(t) \gets$ sense current sensory state
 \State update $e(t)$ based on $\lVert x(t) - p(t)\rVert$
 \If {\textsc{surprise}$(e(t), E(t))$ and not $\textit{searching}$}
 \State $t_s \gets 0$
 \State $\textit{searching} \gets$ true
 \EndIf
 \If {$\textit{searching}$}
 \State $t_s \gets t_s + 1$
 \State $B(t), \textit{searching} \gets$ \textsc{search\_ step}$(t_s)$
 \State $E(t) \gets$ error model associated with $B(t)$
 \Else
 \State update $B(t)$ based on $x(t-1)$, $y(t)$, $x(t)$
 \State update $E(t)$ based on $e(t)$
 \EndIf
 \State \textcolor{black!50}{\Comment{2. action generation and planning}}
 \State $\textit{exploration} \gets$ exploration or planning phase?
 \If {$\textit{exploration}$} \textcolor{black!50}{\Comment{DEP-based exploration}}
 \State update $DEP$ based on $x(t)$ (Eq. \ref{eqCupdate})
 \State $y(t) \gets$ action from $DEP$ given $x(t)$ (Eq. \ref{equCNetwork})
 \Else \textcolor{black!50}{\Comment{goal-directed planning}}
 \If {not $\textit{searching}$}
 \State $x^G \gets$ receive goal state
 \State $B(t) \gets$ \textsc{planning}$(x^G)$
 \If{$B(t) \neq B(t-1)$}
 \State $t_s \gets 0$
 \State $\textit{searching} \gets$ true
 \EndIf
 \EndIf
 \State $y(t) \gets$ action from $B(t)$ given $x(t)$
 \EndIf
 \State \textcolor{black!50}{\Comment{3. next prediction}}
 \State $p(t) \gets $ prediction from $B(t)$ given $x(t)$.
 \State execute action $y(t)$
 \EndWhile
 \EndProcedure
 \end{algorithmic}
 \end{algorithm}

 \begin{algorithm}
 \caption{SUBMODES: model creation} \label{algorithmCreate}
 \begin{algorithmic}[1]
 \Procedure{create\_ new\_ model}{}
 \State $B_i \gets$ create new behavioral model
 \State $E_i \gets$ create new error model
 \For{$B_j \in \mathcal{B}$}
 \State $T_{i \rightarrow j} \gets$ create new transition model
 \State $T_{j \rightarrow i} \gets$ create new transition model
 \EndFor
 \State add created models to $\mathcal{B}, \mathcal{E}$, and $\mathcal{T}$, respectively
 \State \Return $B_i$
 \EndProcedure
 \end{algorithmic}
 \end{algorithm}

 \begin{algorithm}
 \caption{SUBMODES: surprise detection} \label{algorithmSurprise}
 \begin{algorithmic}[1]
 \Procedure{surprise}{$e$, $E_i$}
 \State $\bar{e} \gets$ mean of $E_i$
 \State $\bar{\sigma} \gets$ standard deviation of $E_i$
 \State \Return $e > \bar{e} + \theta \bar{\sigma}$ (Eq. \ref{surprise})
 \EndProcedure
 \end{algorithmic}
 \end{algorithm}

 \begin{algorithm}
 \caption{SUBMODES: one step of searching}
 \begin{algorithmic}[1]
 \Procedure{search\_ step}{$t_s$}
 \For{$B_i \in \mathcal{B}$}
 	\State $\bar{e}_s(i) \gets$ update mean prediction error of $B_i$ \hspace*{2.5cm} during the searching phase
 \EndFor
 \If{$t_s > \tau_{s,min}$} \textcolor{black!50}{\Comment{try to determine next model}}
 	\State $\mathcal{B}' \gets$ set of $B_i \in \mathcal{B}$ with no \textsc{surprise}$(\bar{e}_s(i), E_i)$
 	\If{$\mathcal{B}' \neq \emptyset$} \textcolor{black!50}{\Comment{found suitable models}}
 	\State $B_{next} \gets \argmin_{B_i \in \mathcal{B}'} \bar{e}_s(i)$
 	\If{$B_{next} \neq B(t)$} \textcolor{black!50}{\Comment{model transition}}
 	\State update transition models
 	\EndIf
 	\State \Return $B_{next}$, $\textit{searching} =$ false
 	\ElsIf{$t_s > \tau_{s,max}$} \hspace*{-0.02cm}\textcolor{black!50}{\Comment{no model found in time}}
 	\State $B_{next} \gets$ \textsc{create\_ new\_ model}$()$
 	\State update transition models
 	\State \Return $B_{next}$, $\textit{searching} =$ false
 	\EndIf
 \EndIf
 \State \Return $B(t)$, $\textit{searching} =$ true \textcolor{black!50}{\Comment{continue searching}}
 \EndProcedure
 \end{algorithmic}
 \end{algorithm}

 \begin{algorithm}
 \caption{SUBMODES: planning the next behavior} \label{algorithmPlan}
 \begin{algorithmic}[1]
 \Procedure{planning}{$x^G$}
 \State $B_{next} \gets$ determine best behavioral model for the  \hspace*{2.05cm}
 goal $x^G$ over $\tau_p$ steps (Eq. \ref{eqPlan})
 \State \Return $B_{next}$
 \EndProcedure
 \end{algorithmic}
 \end{algorithm}

\end{document}